\documentclass[journal]{IEEEtran}
\usepackage{graphicx}
\usepackage{subfigure}
\usepackage{epstopdf}
\usepackage{CJK}
\usepackage{amsmath}
\usepackage{breqn}
\usepackage{amsthm}
\usepackage{float}
\usepackage{color}
\usepackage{algorithmic,algorithm}
\usepackage{amsmath}
\usepackage{multirow}
\usepackage{xcolor}
\usepackage{amsfonts}
\usepackage{booktabs}
\usepackage{threeparttable}
\usepackage{cite}

\ifCLASSINFOpdf
\else
\fi

\begin{document}
\title{Detection of Moving Object in Dynamic Background Using Gaussian Max-Pooling and Segmentation Constrained RPCA}

\author{Yang~Li,
        Guangcan~Liu,~\IEEEmembership{Member,~IEEE,}
        Shengyong~Chen,~\IEEEmembership{Senior Member,~IEEE,}

\thanks{Y. Li is with the Department of Computer Science and Engineering, Tianjin University of Technology, 319 Binshuixi Road, Tianjin, China 300384. E-mail: 163102403@stud.tjut.edu.cn.}
\thanks{G. Liu is with B-DAT and CICAEET, School of Information and Control, Nanjing University of Information Science and Technology, 219 Ningliu Road, Nanjing, Jiangsu, China 210044. E-mail: gcliu@nuist.edu.cn. Phone: (86)25-58731276. Fax: (86)25-58731277.}
\thanks{S. Chen is with the Department of Computer Science and Engineering, Tianjin University of Technology, 319 Binshuixi Road, Tianjin, China 300384. E-mail: csy@tjut.edu.cn.}
}
\markboth{Journal of \LaTeX\ Class Files,~Vol.~XX, No.~XX, XXXX}%
{Shell \MakeLowercase{\textit{et al.}}: Bare Demo of IEEEtran.cls for IEEE Journals}

\maketitle

\begin{abstract}
Due to its efficiency and stability, Robust Principal Component Analysis (RPCA) has been emerging as a promising tool for moving object detection. Unfortunately, existing RPCA based methods assume static or quasi-static background, and thereby they may have trouble in coping with the background scenes that exhibit a persistent dynamic behavior. In this work, we shall introduce two techniques to fill in the gap. First, instead of using the raw pixel-value as features that are brittle in the presence of dynamic background, we devise a so-called Gaussian max-pooling operator to estimate a ``stable-value'' for each pixel. Those stable-values are robust to various background changes and can therefore distinguish effectively the foreground objects from the background. Then, to obtain more accurate results, we further propose a Segmentation Constrained RPCA (SC-RPCA) model, which incorporates the temporal and spatial continuity in images into RPCA. The inference process of SC-RPCA is a group sparsity constrained nuclear norm minimization problem, which is convex and easy to solve. Experimental results on seven videos from the CDCNET 2014 database show the superior performance of the proposed method.
\end{abstract}


\IEEEpeerreviewmaketitle

\section{Introduction}
\IEEEPARstart{I}{n} the areas of high safety standards such as stations, airports, schools, banks and so on, surveillance cameras are now ubiquitous, producing a large number of videos every day. The massive nature of surveillance videos makes it very difficult for human investigators to manually search a target through all videos. Therefore, it is urgent to enable the surveillance system to intelligently detect irregularities, suspicious targets, etc. To this end, moving object detection (or background subtraction), which aims to find independent moving objects in a scene, is an important preprocessing step. Many methods have been proposed and investigated in the literature over the past several years, e.g.,~\cite{paper01,paper02,paper03,paper04,paper05,paper06,paper07,paper08,paper09,paper10,paper11,paper12,paper13,paper14}.
\begin{figure}[!t]
\centering
\subfigure[]{
\begin{minipage}{4cm}
\centering
\includegraphics[height=3cm, width=4cm]{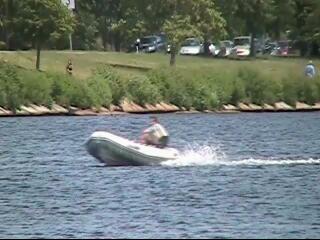}
\end{minipage}
}
\subfigure[]{
\begin{minipage}{4cm}
\centering
\includegraphics[height=3cm, width=4cm]{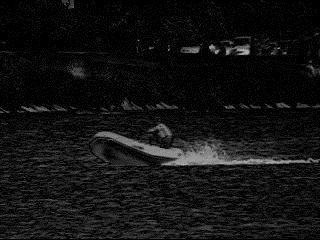}
\end{minipage}
}

\caption{Exemplifying the performance of RPCA in dealing with dynamic background: (a) an input frame with ``surface wave''; (b) detected foreground (white areas). In this example, many background areas are mistaken for foreground moving objects.}\label{fig:graph:rpca}
\end{figure}

While the background of the scene is fixed and static, there are strong correlations between the video frames. In this case, Robust Principal Component Analysis (RPCA)~\cite{Candes:2009:JournalACM} has already provides us a convenient way to perform moving object detection. To be more precise, one could stack each frame as a column of a matrix $D\in\mathbb{R}^{m\times{}n}$ at first, then decompose $D$ into a low-rank term and a sparse term by solving the following convex problem:
\begin{align}\label{eq:rpca}
  \min_{A, E} & \|A\|_*+\lambda\|E\|_1,\textrm{ s.t. }D = A + E,
\end{align}
where $\|\cdot\|_*$ is the nuclear norm of a matrix, $\|\cdot\|_1$ denotes the $\ell_1$ norm of a matrix seen as a long vector, and $\lambda>0$ is a parameter. As shown in~\cite{Candes:2009:JournalACM}, the low-rank and sparse components correspond to the static background and the moving objects, respectively. However,  RPCA relies heavily on the assumption that the background is static or quasi-static and is therefore not applicable to the realistic tasks with dynamic background (see Figure~\ref{fig:graph:rpca}).

Some researchers have tried to overcome the drawbacks of RPCA by making use of some additional priors. Zhou et al.~\cite{paper03} considered a prior that the foreground objects are contiguous pieces with relatively small size. This method does improve the identifiability of the foreground objects but still replies on the assumption that the background is quasi-static. Cui et al.~\cite{paperCui} extended the RPCA model to process the videos captured by moving cameras. This method works well when the camera motion is predictable, but it cannot handle well the cases where the background scenes exhibit a persistent dynamic behavior, the motion of which is essentially unpredictable. Overall, it is still not enough for existing RPCA based methods
to cope with the moving object detection problem in the context of dynamic background.
\begin{figure}[!t]
\centering
\includegraphics[width=8.5cm]{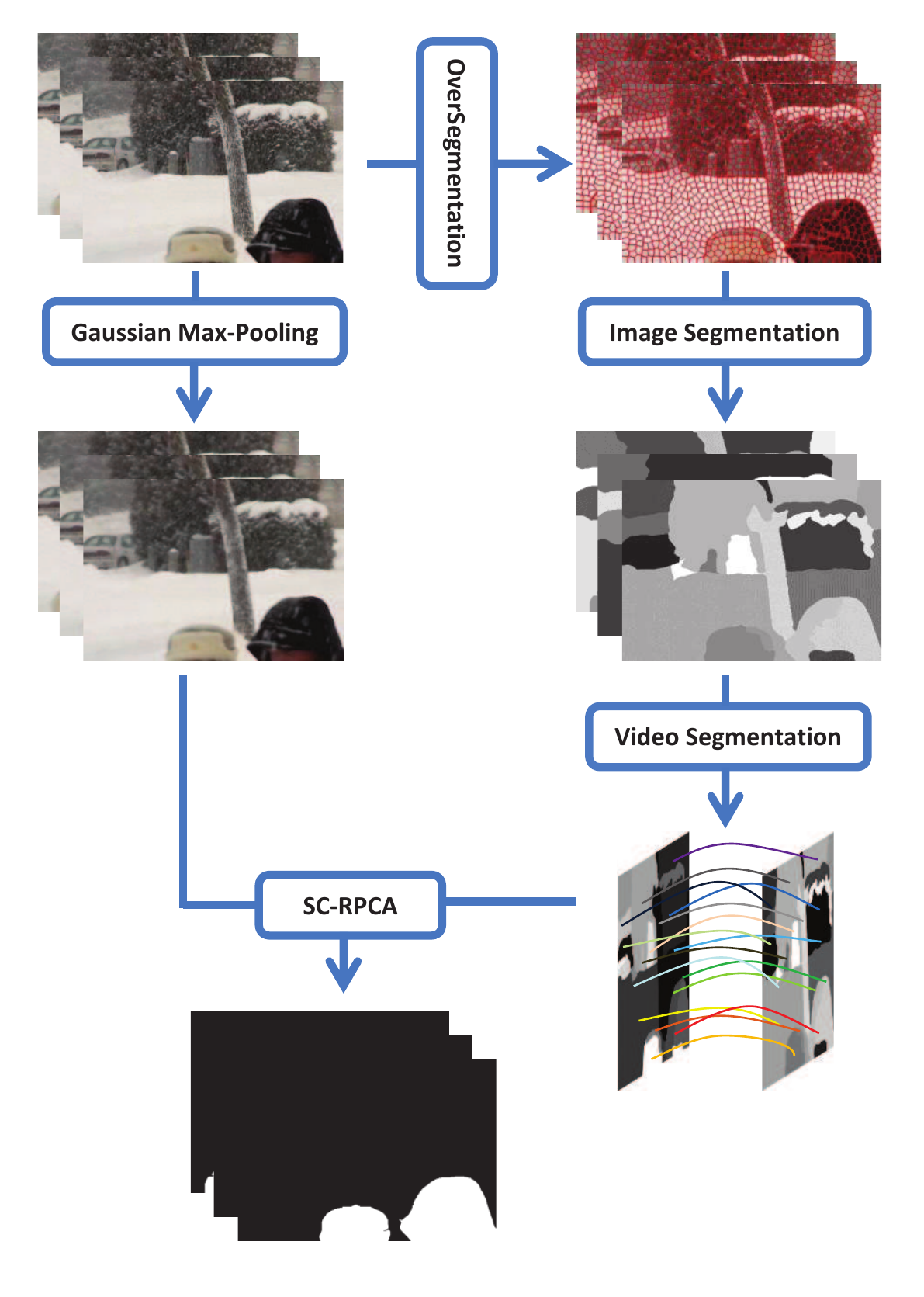}
\caption{Outline of the proposed method for moving object detection.}
\label{fig:graph:ourmethod}
\end{figure}

In this work, we introduce a novel RPCA based method for detecting moving objects in the scenes with dynamic background. As shown in Figure~\ref{fig:graph:ourmethod}, our method contains two key techniques: 1) Instead of using the raw pixels as features that are brittle while dealing with dynamic background, we estimate a ``stable-value'' for each pixel by a novel operator called Gaussian max-pooling. Roughly speaking, Gaussian max-pooling is an elegant technique for selecting the most frequent value from a collection of candidate values. Equipped with those stable-values that sever as good features for distinguishing the foreground objects from the background, RPCA could produce moderately good results by taking as inputs the estimated stable-values rather than the raw pixels. 2) To achieve more accurate results, we shall take into account the temporal and spatial continuity in images.
The temporal and spatial continuity are obtained by first oversegmenting the images into superpixels~\cite{Mori:2005:GMS}, then using the subspace clustering algorithm proposed in~\cite{Ma:2007:SMM} to group the superpixels into much larger subregions (image segmentation), and finally using the algorithm in~\cite{Ma:2007:SMM} again to group together the similar subregions from different image frames (video segmentation). The segmentation result are considered as a group sparsity constraint and incorporated into RPCA, resulting in a novel model termed Segmentation Constrained RPCA (SC-RPCA). The inference process of SC-RPCA is a group sparsity constrained nuclear norm minimization problem, which is convex and easy to solve. Experimental results on seven videos from the CDCNET 2014 database show the superior performance of the proposed method. In summary, the contributions of this paper include:
\begin{itemize}
\item[$\diamond$] We devise a novel operator called Gaussian max-pooling to estimate the stable-value for each pixel. The stable-values are robust to various background variations such as illumination change, waves in water bodies, falling snow, and so on.
\item[$\diamond$] We proposed a novel model termed SC-RPCA, which integrates low-rankness and group sparsity into a unified framework. Different from the existing methods that use the $\ell_{2,1}$ norm to characterize the group sparsity prior~\cite{xu:2012:tit,liu:tpami:2016, paperCui}, SC-RPCA utilizes a generalized $\ell_{2,1}$ to cope with more complicate group sparsity constraints.
\item[$\diamond$] Based on the proposed Gaussian max-pooling and SC-RPCA, we establish a novel method for moving object detection in dynamic background. Experimental results show that our method is quite competitive while comparing to the state-of-the-art methods.
\end{itemize}

The remaining of this paper is organized as follows. Section~\ref{sec:rw} summarizes some related works. Section~\ref{sec:proposed} presents the details of the proposed method. Section~\ref{sec:exp} show empirical results and Section~\ref{sec:con} concludes this paper.
\section{Related Work}\label{sec:rw}
Due to the correlation between frames, video is an attractive place for low-rank subspaces. As a subspace can be well modeled by a (degenerate) Gaussian distribution~\cite{tpami_2013_lrr}, it is natural to consider the Gaussian model as a candidate for background modeling. Wren et al.~\cite{paper05} proposed a single Gaussian model to model the background in video frames. Their method procures preferable effects on indoor scenes, but it cannot effectively detect objects in the outdoor which is often a multi-modal environment. Stauffer et al.~\cite{paper06} established an object detection method based on using Gaussian Mixture Model (GMM)~\cite{paperGMM} to model the background. The processing of each pixel is independent of each other and each pixel is composed of multiple weights of different Gaussian mixture of superposition of distribution. Unlike the single Gaussian model, this method fully uses the historical information to represent the background and therefore can adapt to multi-modal environments. However, its computational cost is too large to promptly deal with sudden background changes. Splitting Gaussian in Mixture Models (SGMM)~\cite{paper07} uses two complementary hybrid Gaussian models with different update rates: One is used to detect moving objects precisely. The other is used to build the background. Comparing to GMM, SGMM can not only improve detection accuracy but also reduce the computational complexity.

Elgammal et al.~\cite{paper08} adopted a Kernel Density Estimation (KDE) for background modeling. The probability density distribution of each pixel is estimated by apply the Gaussian kernel filter to the continuous pixels in all frames. Those densities serve as the basis for judging whether a pixel belongs to a moving object. Generally, this method has strong adaptability and high accuracy. Visual Background Extractor (ViBE)~\cite{paper09} supposes that each pixel has similar distributions as its domain pixels in the spatial domain. The method uses domain pixels to build the background model, which is compared with the currently input pixel values to determine the foreground objects. Pixel-Based Adaptive Segmenter (PBAS)~\cite{paper04} is a non-parametric model based on the gradients of front pixels. It could be robust to slow illumination changes in the background. However, as the spatial continuity is not well considered, this method may fail when dealing with a background of persistent dynamic behavior.

Wang et al.~\cite{paper10} proposed a hybrid, multi-level method termed FTSG, which combines Flux Tensor-based motion detection with the classification results from a Split Gaussian mixture model. Sedky et al.~\cite{paper11} presented a change detection technique based on the dichromatic color reflectance model. Lu et al.~\cite{paper12} used a multi-scale background model for motion detection by following a nonparametric paradigm: Each location in a dynamic scene consisting of a set of samples on different spatial scales. Liang et al.~\cite{paper13} proposed an online object detection method that is robust to sudden illumination changes and regular dynamic background. Recently, Gregorio et al.~\cite{paper14} leveraged Weightless Neural Networks (WNN) to perform moving object detection in dynamic background.

\section{Our Method for Detecting Moving Object in Dynamic Background}\label{sec:proposed}
This section details the proposed techniques, mainly including the approach of Gaussian max-pooling and the model of Segmentation Constrained RPCA (SC-RPCA).

\subsection{Gaussian Max-Pooling}
Given a sequence of images $\{I_1,I_2,\cdots,I_N\}$, one could form a data matrix $D$ by stacking each image as a column of $D$ at first, then decompose $D$ into a low-rank term $A$ and a sparse term $E$ by solving the RPCA problem in~\eqref{eq:rpca}. In the ideal case where the background is strictly static, the term $A$ will be a rank-1 matrix and the nonzero values in $E$ can exactly identify anything that moves. This ``strength'', however, by itself may become a weakness in the context of dynamic background. As shown in Figure~\ref{fig:graph:rpca}, RPCA may wrongly judge many background areas as foreground whenever the background is exhibiting a persistent dynamic behavior.
\begin{table}
\caption{Examples of the sampling windows.}\label{tab:sw}
\centering
\subtable[sample 1]{
\begin{tabular}{|c|c|c|c|c|}\hline
    46 & 53 & 50 & 51 & 68 \\ \hline
    68 & 68 & 53 & 45 & 49 \\ \hline
    63 & 62 & {\bf 58} & 62 & 48 \\ \hline
    59 & 52 & 43 & 47 & 59 \\ \hline
    82 & 79 & 65 & 62 & 45 \\ \hline
\end{tabular}
}
\qquad
\subtable[sample 2]{
\begin{tabular}{|c|c|c|c|c|}\hline
    46 & 53 & 63 & 62 & 59 \\ \hline
    68 & 69 & 52 & 82 & 79 \\ \hline
    128 & 128 & {\bf 128} & 128 & 128 \\ \hline
    128 & 128 & 128 & 128 & 128 \\ \hline
    128 & 128 & 128 & 128 & 128 \\ \hline
\end{tabular}

}
\end{table}

In order to relieve the drawbacks of RPCA, it is straightforward to exact some features that are robust to background variations. To do this, we would like to consider a simple and effective approach: For a certain pixel $v$, consider its neighbor pixels in the spatial domain, denoted as $\mathcal{N}_v\{v_1,v_2,\cdots,v_{n^2}\}$, where $n$ is an odd number indicates the window size of sampling. In this work, the sampling window for a pixel is simply a square image region around the pixel, as shown in Table~\ref{tab:sw}. Given $\mathcal{N}_v$, the stable-value $p_v$ corresponding to the pixel $v$ could be computed as the pixel value that appears most frequently in $\mathcal{N}_v$, i.e.,
\begin{align*}
p_v = \arg\max_{u\in\mathcal{N}_v}\mathrm{Pr}(u).
\end{align*}
While simple and straightforward, it is indeed not easy to estimate the priori probability of each pixel accurately, as the sampling window is often small (e.g., $5\times5$). Therefore, we shall propose a Gaussian max-pooling operator to estimate the stable-value of each pixel. Denote by $\mathcal{M}$ all the possible pixel values in image domain, i.e., $\mathcal{M}=\{0,1,\cdots,255\}$. For a certain pixel $v$, we compute its stable-value by maximizing a posterior probability:
\begin{align}\label{eq:stable-map}
p_v = \arg\max_{u\in\mathcal{M}}\mathrm{Pr}(\mathcal{N}_v|u)=\arg\max_{u\in\mathcal{M}}\sum_{u'\in\mathcal{N}_v}\mathrm{Pr}(u'|u).
\end{align}
As usual, we would assume that the conditional distribution is Gaussian and, accordingly, estimate the conditional probability of a pixel $u'$ given another pixel $u$ by:
\begin{align}\label{eq:condp}
 \mathrm{Pr}(u'|u)=\frac{1}{\sqrt{2\pi }\sigma }\exp(-\frac{(u'-u)^2}{2\sigma ^2}),
\end{align}
where $\sigma>0$ is taken as a parameter.
\begin{algorithm}[h]
  \caption{Gaussian Max-Pooling}\label{alg:gmp}
  \begin{algorithmic}[1]
    \REQUIRE
    an image $I$.
    \ENSURE
      the processed image $Ip$.
    \STATE Parameters: size of sampling window and $\sigma$.
    \FOR{each pixel $v$ in $I$} 
         \STATE construct a window $\mathcal{N}_v$ around the pixel $v$.
         \STATE compute the conditional probabilities by~\eqref{eq:condp}.
         \STATE compute the stable-value of $v$ by~\eqref{eq:stable-map}.
    \ENDFOR
    \STATE form a new image $Ip$ by replacing the pixel-values in $I$ with their respective stable-values.
    \RETURN $Ip$
  \end{algorithmic}
\end{algorithm}

\begin{figure}[!t]
\centering
\subfigure[]{
\begin{minipage}{4cm}
\centering
\includegraphics[height=3cm, width=4cm]{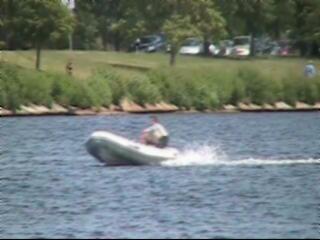}
\end{minipage}
}
\subfigure[]{
\begin{minipage}{4cm}
\centering
\includegraphics[height=3cm, width=4cm]{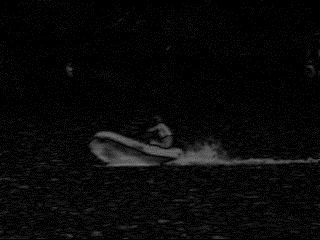}
\end{minipage}
}
\caption{Exemplifying the effects of Gaussian max-pooling: (a) the input frame shown in Figure~\ref{fig:graph:rpca}(a); (b) detection results produced by using the stable-values rather than the pixel-values as inputs for RPCA.}
\label{fig:graph:gmp}
\end{figure}

Algorithm~\ref{alg:gmp} summarizes the computational procedures of the Gaussian max-pooling operator. We also give some empirical results in Figure~\ref{fig:graph:gmp}. It can be seen that the detection results shown in Figure~\ref{fig:graph:gmp}(b) are much better than Figure~\ref{fig:graph:rpca}(b). Namely, the background mistaken for foreground in Figure~\ref{fig:graph:gmp}(b) is less than Figure~\ref{fig:graph:rpca}(b). This demonstrates the effectiveness of our Gaussian max-pooling.

\subsection{Segmentation Constrained RPCA}
While better than Figure~\ref{fig:graph:rpca}(b), the results in Figure~\ref{fig:graph:gmp}(b) are still far from perfect, as there are still many background pixels mistaken for foreground. To obtain more accurate detection, we shall taken into account the spatial and temporal continuity widely existing in videos.

\begin{figure}[!t]
\centering
\subfigure[]{
\begin{minipage}{4cm}
\centering
\includegraphics[height=3cm, width=4cm]{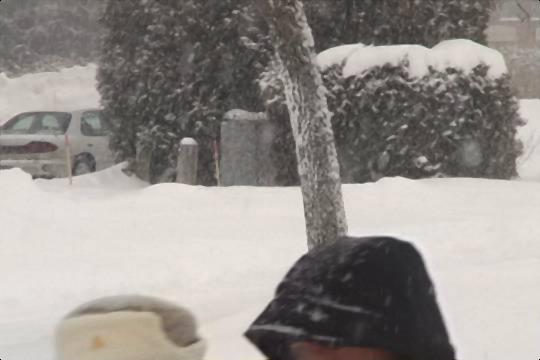}
\end{minipage}
}
\subfigure[]{
\begin{minipage}{4cm}
\centering
\includegraphics[height=3cm, width=4cm]{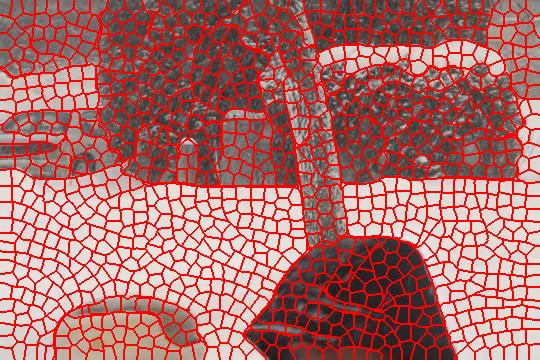}
\end{minipage}
}
\subfigure[]{
\begin{minipage}{4cm}
\centering
\includegraphics[height=3cm, width=4cm]{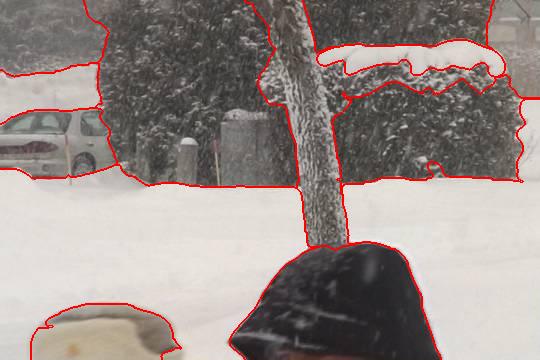}
\end{minipage}
}
\subfigure[]{
\begin{minipage}{4cm}
\centering
\includegraphics[height=3cm, width=4cm]{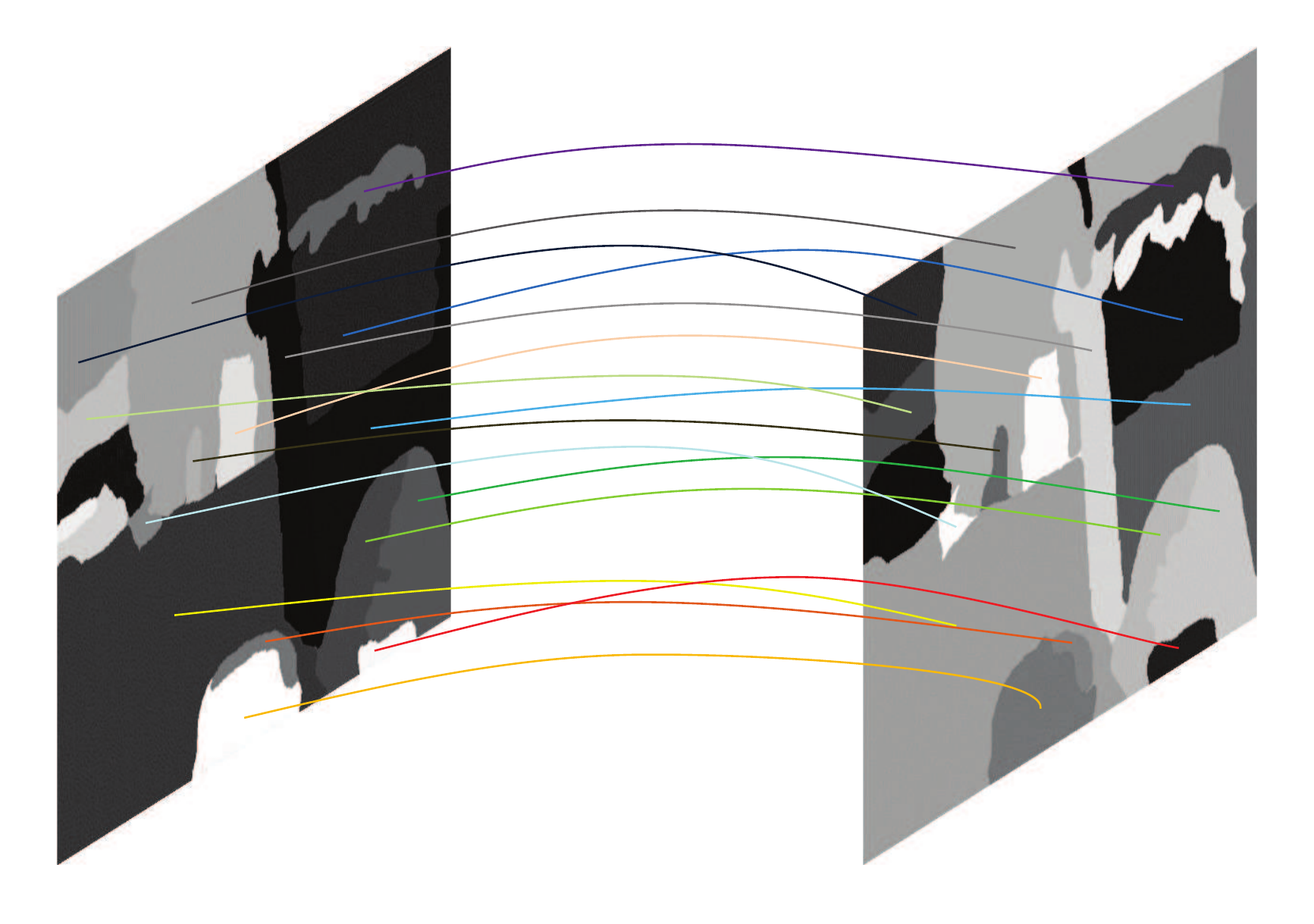}
\end{minipage}
}
\caption{Obtaining the spatial and temporal continuity in videos: (a) an image; (b) superpixels; (c) results of image segmentation; (d) merging together similar subregions across different frames.}
\label{fig:graph:st}
\end{figure}

\subsubsection{Acquiring the spatial and temporal continuity in videos} The desired continuity information is obtained by grouping together the pixels that are similar in appearance and adjacent in the space or time domain. First, we use the approach established in~\cite{Mori:2005:GMS} to oversegment the images into superpixels (Figure~\ref{fig:graph:st}(b)). Then we use the subspace clustering algorithm proposed in~\cite{Ma:2007:SMM} to group the superpixels into much larger subregions (Figure~\ref{fig:graph:st}(c)); this procedure is also known as image segmentation. Finally, we use again the algorithm in~\cite{Ma:2007:SMM} to group together the similar, adjacent subregions across different frames (Figure~\ref{fig:graph:st}(d)). Here, two subregions are considered to be adjacent to each other if and only if: 1) their corresponding frames are adjacent and 2) their regional center distance is smaller than a certain threshold. The whole procedure is called video segmentation and summarized in Algorithm~\ref{alg:vs}.
\begin{algorithm}[h]
  \caption{Video Segmentation}
  \label{alg:vs}
  \begin{algorithmic}[1]
    \REQUIRE
      a sequence of image frames $\{I_1,\cdots,I_N\}$.
    \ENSURE
    segmentation results
    \FOR{$i=1,\cdots,N$}
    \STATE oversegment $I_i$ into superpixels by~\cite{Mori:2005:GMS}.
    \STATE cluster the superpixles of $I_i$ into subregions by~\cite{Ma:2007:SMM}.
    \ENDFOR
    \REPEAT
    \STATE find all the adjacent subregions across all $N$ frames.
    \STATE group together the most similar subregions according to the criterion proposed in~\cite{Ma:2007:SMM}.
    \UNTIL {no subregions can be grouped}
  \end{algorithmic}
\end{algorithm}

\subsubsection{The model of SC-RPCA} Up to now, we have obtained a segmentation for a video (i.e., a sequence of images). For the ease of discussion, we denote the segmentation as
\begin{align*}
C = C_1\cup{}C_2\cup\cdots\cup{}C_m,
\end{align*}
where $m$ is the number of groups automatically determined by the algorithm in~\cite{Ma:2007:SMM}. For the image pixels belonging to the same group, it is natural to expect that their foregorund/backgournd labels are the same. To this end, we incorporate the constraints encoded by $C$ into RPCA, resulting a novel model termed SC-RPCA:
\begin{eqnarray}\label{eq:sc-rpca}
  \min_{A,E} & \|A\|_*+\lambda\|E\|_{C(2,1)},&\textrm{ s.t. }D_P = A + E,
\end{eqnarray}
where $D_p$ is $D$ processed by Gaussian max-pooling and $\|\cdot\|_{C(2,1)}$ is the generalized $\ell_{2,1}$ norm associated with a segmentation $C$. More precisely, the generalized $\ell_{2,1}$ norm is defined by
\begin{eqnarray}\label{eq:gl21}
\|E\|_{C(2,1)} = \sum_{i=1}^{m}\sqrt{|C_i|\sum_{(j,k)\in{}C_i}([E]_{jk})^2},
\end{eqnarray}
where $[\cdot]_{ij}$ is the $(i,j)$th entry of a matrix, and $|C_i|$ denotes the number of pixels in the group $C_i$. It is easy to see that, whenever each column in $D_p$ forms a group, $\|\cdot\|_{C(2,1)}$ becomes the traditional $\ell_{2,1}$ norm and therefore SC-RPCA falls back to the RPCA model analyzed in~\cite{xu:2012:tit}.

\begin{figure}[!t]
\centering
\subfigure[]{
\begin{minipage}{4cm}
\centering
\includegraphics[height=3cm, width=4cm]{1-2.jpg}
\end{minipage}
}
\subfigure[]{
\begin{minipage}{4cm}
\centering
\includegraphics[height=3cm, width=4cm]{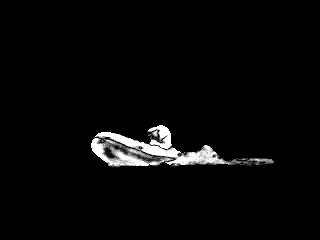}
\end{minipage}
}
\caption{Exemplifying the effects of SC-RPCA: (a) the input frame shown in Figure~\ref{fig:graph:rpca}(a); (b) detection results produced by SC-RPCA, using the stable-values computed by Gaussian max-pooling as features.}
\label{fig:graph:gcrpca}
\end{figure}

Thanks to the effects of the low-rankness and group sparsity constraints, SC-RPCA can seamlessly integrate various priors into a unified, convex procedure, including the background correlation, the foreground sparsity, and the temporal/spatial continuity. In additional, the $\ell_{C(2,1)}$ norm is also tolerant to the small mistakes made in the video segmentation procedure. Figure~\ref{fig:graph:gcrpca} shows an example, which demonstrates the strengths of SC-RPCA.
\subsubsection{Optimization algorithm}
The problem in~\eqref{eq:sc-rpca} is convex and can be optimized efficiently with the ALM method~\cite{lin_alm}, which minimizes the following augmented Lagrange function:
\begin{dmath}
\mathcal{L}(A,E,Y,\mu)=\left \| A \right \|_{*}+\lambda \left \| E \right \|_{C(2,1)}+\left \langle Y,D_P-A-E \right \rangle+\frac{\mu }{2}\left \| D_P-A-E \right \|_{2}^{F},
\end{dmath}
where $Y$ is Lagrange multipliers and $\mu$ is a penalty parameter. The inexact ALM method, which is also called the alternating direction method (ADM), is outlined in Algorithm~\ref{alg:ialm}. Notice that the subproblems of the
algorithm are convex and have closed-form solutions. More precisely, Step 3 is solved via the singular value thresholding operator~\cite{svt:cai:2008}, whereas Step 4 is solved as follows:
\begin{align*}
&E=\arg\min_{E}\mathcal{L}(A,E,Y,\mu) \\
&= \arg\min_{E}\frac{\lambda}{\mu} \|E\|_{C(2,1)}+\frac{1}{2}\left \| E-\left ( D_P-A+\frac{Y}{\mu} \right ) \right \|_{2}^{F}.
\end{align*}
Write $M=D_P-A+Y/\mu$, assume that $(j,k)\in{}C_i$ without loss of generality, and denote $\|[M]_{C_i}\|_2=\sqrt{\sum_{(a,b)\in{}C_i}([M]_{ab})^2}$. Then the solution to the above problem is given by
\begin{align*}
&[E]_{jk} = \left\{
\begin{array}{cc}
\frac{\|[M]_{C_i}\|_2-\frac{\lambda{}|C_i|}{\mu}}{\|[M]_{C_i}\|_2}[M]_{jk}, & \textrm{if }\|[M]_{C_i}\|_2>\frac{\lambda{}|C_i|}{\mu},\\
0,& \textrm{otherwise,}
\end{array}
\right.
\end{align*}
which follows from Lemma 3.2 of~\cite{icml_2010_lrr}.
\begin{algorithm}[h]
  \caption{Solving Problem~\eqref{eq:sc-rpca} by Inexact ALM}\label{alg:ialm}
  \begin{algorithmic}[1]
    \REQUIRE
      feature matrix $D_P$ and segmentation $C$.
    \ENSURE
      the optimal solution $(A, E)$
    \STATE initialization: $A=E=Y=0, \mu =10^{-6}$.
    \WHILE{not converged}
     \STATE fix the others and update $A$ by
     $$A=\arg\min_{A}\mathcal{L}(A,E,Y,\mu).$$
     \STATE fix the others and update $E$ by
     $$E=\arg\min_{E}\mathcal{L}(A,E,Y,\mu).$$
     \STATE update the multipliers by
     $$Y=Y+\mu(D_P-A-E).$$
     \STATE update the penalty parameter
     $$\mu = \rho\mu,$$
     where the parameter $\rho$ takes the role of controlling the convergence speed. It is set as $\rho=1.1$ in all experiments.
    \ENDWHILE
    \RETURN $(A , E)$
  \end{algorithmic}
\end{algorithm}
\subsubsection{Overall algorithm for moving object detection}
With the optimal solution $(A, E)$ to~\eqref{eq:sc-rpca}, all the image frames are segmented to foreground and background simultaneously. Namely, the locations of the nonzero elements in $E$ indicate the locations of foreground objects. Algorithm~\ref{alg:mod} summarizes the whole procedure of our method for detecting moving objects in videos.
\begin{algorithm}[h]
  \caption{Moving Object Detection}\label{alg:mod}
  \begin{algorithmic}[1]
    \REQUIRE
      a video consisting of $N$ images $\{I_1,\cdots,I_N\}$.
    \ENSURE
      the detection results.
    \STATE process all the images by Algorithm~\ref{alg:gmp}, resulting in a sequence of processed images $\{Ip_1,\cdots,Ip_N\}$.
    \STATE obtain a video segmentation $C$ by Algorithm~\ref{alg:vs}.
    \STATE form a matrix $D_P$ by stacking each processed image as a column of the matrix.
    \STATE obtain a sparse matrix $E$ by Algorithm~\ref{alg:ialm}, using $D_P$ and $C$ as inputs.
    \STATE obtain the final results according to $E$.
  \end{algorithmic}
\end{algorithm}

\section{Experiments and analysis}\label{sec:exp}
\subsection{Experimental Setting}
\subsubsection{Data}
\begin{figure}[t]
\centering
\subfigure[]{
\begin{minipage}{4cm}
\centering
\includegraphics[ width=4cm,height=3cm]{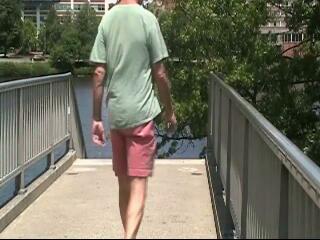}
\end{minipage}
}
\subfigure[]{
\begin{minipage}{4cm}
\centering
\includegraphics[ width=4cm,height=3cm]{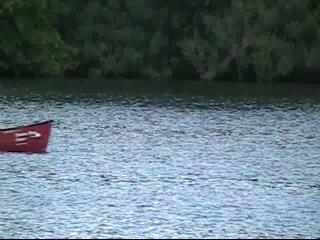}
\end{minipage}
}
\subfigure[]{
\begin{minipage}{4cm}
\centering
\includegraphics[ width=4cm,height=3cm]{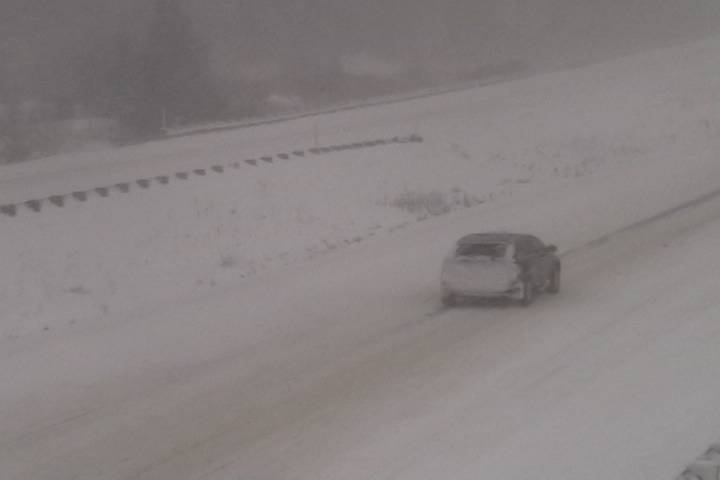}
\end{minipage}
}
\subfigure[]{
\begin{minipage}{4cm}
\centering
\includegraphics[ width=4cm,height=3cm]{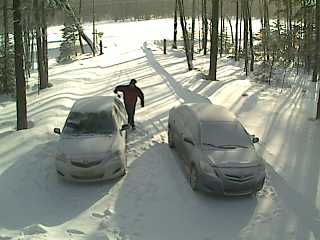}
\end{minipage}
}
\caption{Some challenging examples from the CDCNET 2014 database: (a) overpass video. (b) canoe video. (c) snowfall video. (d) winterDriveway video.}
\label{fig:graph:exam}
\end{figure}

In order to verify the validity of our method, we experiment with seven videos from the CDCNET 2014 database, where the human-annotated ground truth is available. Those videos contain challenging examples in motion detection, e.g.,  dynamic background (waving trees, surface waves), bad weather (snowfall), intermittent object motion, etc. Figure~\ref{fig:graph:exam} shows some challenging examples. Note that there are waving trees in overpass video that cause the background to change. In Figure~\ref{fig:graph:exam}(c), the snowfall affects the integrity of moving object detection. Regarding the example in Figure~\ref{fig:graph:exam}(d), the intermittent object motion makes the pixels of moving object unchanged in several consecutive frames .

\subsubsection{Evaluation metrics}
The following three measures are used for evaluation: 1) True Positive ($TP$), which denotes the number of true foreground pixels correctly classified as foreground. 2) True Negative ($TN$), which is the number of foreground pixles wrongly classified as background. 3) False Positive ($FP$), which is the number of background pixels wrongly classified as foreground. Then, Recall, Precision and F-measure are used to evaluate various moving object detection algorithms:
\begin{eqnarray}
Recall=\frac{TP}{TP+TN}
\end{eqnarray}
\begin{eqnarray}
Precision=\frac{TP}{TP+FP}
\end{eqnarray}
\begin{eqnarray}
F-measure=2\frac{Recall*Precision}{Recall+Precision}
\end{eqnarray}
\subsubsection{Baselines}
To verify the effectiveness of the proposed Gaussian max-pooling operator and the SC-RPCA model, we include for comparison the RPCA based methods, including ``RPCA$+$pixels'' (which uses raw pixels as inputs for RPCA) and ``RPCA$+$stable-value'' (which takes the state-values estimated by our Gaussian max-pooling as inputs). To show the superiority of our moving object detection method, we also include for comparison several state-of-the-art methods, including XUE~\cite{paper17}, Euclidean distance (ED)~\cite{paperEuclideandistance}, MST~\cite{paper12}, Spectral~\cite{paper11}, FSTG~\cite{paper10}, CwisarDH~\cite{paper14}, GMM~\cite{paperGMM}, CP3online~\cite{paper13}, BinWang~\cite{paperBinWang} and KDE~\cite{paper08}.
\subsection{Results}
\subsubsection{Comparison with RPCA based methods}
 Table~\ref{tab:performance_comparison:rpca} shows the evaluation results on the snowfall video. It is can be seen that the original approach of RPCA+pixel is largely outperformed by RPCA+stable-value, which is further distinctly outperformed by SC-RPCA+stable-value. These results illustrate that both the Gaussian max-pooling operator and the SC-RPCA model are effective in detecting moving objects in dynamic background.
\begin{table}[th]
\setlength\tabcolsep{0.05in}
  \centering
  \begin{threeparttable}
  \caption{Comparison results on the snowfall video.}
  \label{tab:performance_comparison:rpca}
    \begin{tabular}{cccc}
    \toprule
    Measure&RPCA+pixel &RPCA+stable-value &SC-RPCA+stable-value\cr
    \midrule

        Recall     &0.128 &0.361 &0.957\cr
                                         Precision  &0.949 &0.531 &0.787\cr
                                         F-measure  &0.226 &0.430 &0.863\cr
    \bottomrule
    \end{tabular}
    \end{threeparttable}
\end{table}

\subsubsection{Comparison with state-of-the-art methods}
\renewcommand\arraystretch{1.2}
\begin{table*}[th]
\setlength\tabcolsep{0.1in}
  \centering
  \begin{threeparttable}
  \caption{Evaluation results on seven videos from the CDCNET 2014 database.}
  \label{tab:performance_comparison}
    \begin{tabular}{ccccccccccccc}
    \toprule

    Video&Measure&\cite{paper17} &\cite{paperEuclideandistance} &\cite{paper12} &\cite{paper11} &\cite{paper10} &\cite{paper14}  &\cite{paperGMM} &\cite{paper13}  &\cite{paperBinWang} &\cite{paper08} &proposed\cr

    \midrule

    \multirow{3}{*}{snowfall}            &Recall    &0.360 &0.993 &0.990 &0.637 &0.699 &0.859 &0.630 &0.640 &0.986 &0.986 &0.955\cr
                                         &Precision &0.802 &0.358 &0.496 &0.976 &0.982 &0.844 &0.955 &0.978 &0.592 &0.613 &0.930\cr
                                         &F-measure &0.497 &0.529 &0.665 &0.758 &0.812 &0.843 &0.751 &0.763 &0.738 &0.754 &\textbf{0.942}\cr
    \cr
    \multirow{3}{*}{blizzard}            &Recall    &0.693 &1.000 &0.510 &0.753 &0.943 &0.997 &0.915 &0.972 &0.998 &0.995 &0.933\cr
                                         &Precision &0.875 &0.247 &0.999 &0.962 &0.969 &0.491 &0.930 &0.901 &0.618 &0.686 &0.979\cr
                                         &F-measure &0.769 &0.385 &0.661 &0.817 &0.954 &0.652 &0.922 &0.935 &0.756 &0.806 &\textbf{0.955}\cr
    \cr
    \multirow{3}{*}{skating}             &Recall    &0.562 &1.000 &0.670 &0.938 &0.991 &0.993 &0.984 &0.962 &0.997 &0.985 &0.995\cr
                                         &Precision &0.426 &0.558 &0.948 &0.913 &0.960 &0.978 &0.951 &0.966 &0.968 &0.972 &0.947\cr
                                         &F-measure &0.478 &0.681 &0.778 &0.925 &0.975 &\textbf{0.986} &0.967 &0.964 &0.982 &0.979 &0.970\cr
    \cr
    \multirow{3}{*}{Boats}               &Recall    &0.845 &0.811 &0.806 &0.702 &0.830 &0.875 &0.543 &0.730 &0.730 &0.682 &0.957\cr
                                         &Precision &0.793 &0.718 &0.922 &0.991 &0.941 &0.865 &0.930 &0.757 &0.919 &0.957 &0.787\cr
                                         &F-measure &0.818 &0.761 &0.859 &0.820 &\textbf{0.880} &0.864 &0.685 &0.730 &0.813 &0.796 &0.863\cr
    \cr
    \multirow{3}{*}{canoe}               &Recall    &0.516 &0.840 &0.708 &0.626 &0.838 &0.749 &0.638 &0.767 &0.957 &0.672 &0.865\cr
                                         &Precision &0.308 &0.763 &0.937 &0.994 &0.966 &0.989 &0.994 &0.956 &0.859 &0.951 &0.944\cr
                                         &F-measure &0.387 &0.792 &0.805 &0.767 &0.896 &0.852 &0.776 &0.851 &\textbf{0.905} &0.786 &0.901\cr
    \cr
    \multirow{3}{*}{overpass}            &Recall    &0.424 &0.988 &0.700 &0.555 &0.842 &0.690 &0.748 &0.571 &0.952 &0.656 &0.886\cr
                                         &Precision &0.819 &0.624 &0.992 &0.980 &0.982 &0.993 &0.769 &0.996 &0.917 &0.980 &0.830\cr
                                         &F-measure &0.559 &0.764 &0.820 &0.704 &0.906 &0.815 &0.759 &0.723 &\textbf{0.934} &0.786 &0.851\cr
    \cr
    \multirow{3}{*}{winterDriveway}      &Recall    &0.481 &1.000 &0.951 &0.572 &0.895 &0.917 &0.931 &0.729 &0.839 &0.877 &0.734\cr
                                         &Precision &0.917 &0.561 &0.755 &0.998 &0.932 &0.593 &0.794 &0.926 &0.881 &0.849 &0.993\cr
                                         &F-measure &0.630 &0.714 &0.840 &0.727 &\textbf{0.913} &0.711 &0.857 &0.814 &0.858 &0.862 &0.843\cr

    \bottomrule
    \end{tabular}
    \end{threeparttable}
\end{table*}

\begin{figure*}[htbp]
\centering
\subfigure{
\begin{minipage}[b]{0.13\textwidth}
\includegraphics[width=2.3cm,height=1.7cm]{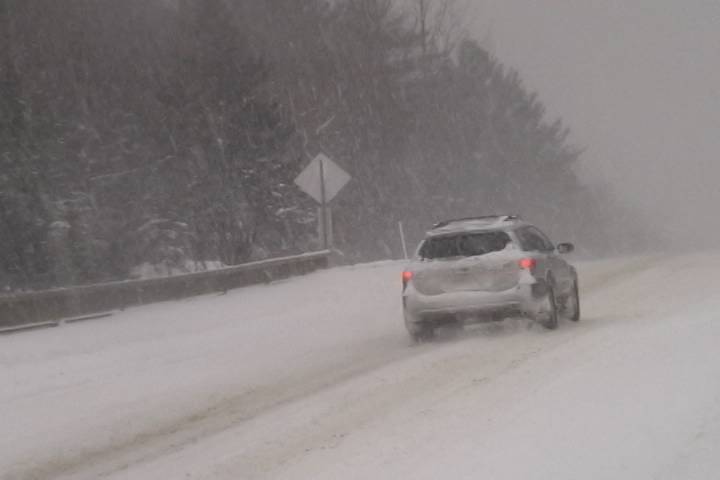}
\end{minipage}
\begin{minipage}[b]{0.13\textwidth}
\includegraphics[width=2.3cm,height=1.7cm]{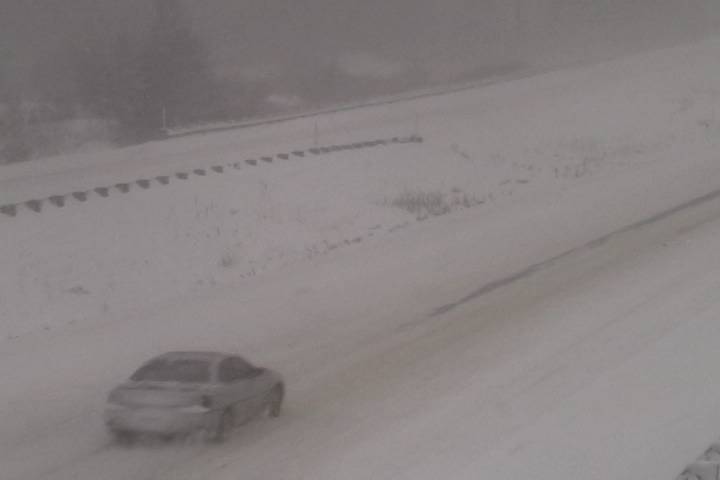}
\end{minipage}
\begin{minipage}[b]{0.13\textwidth}
\includegraphics[width=2.3cm,height=1.7cm]{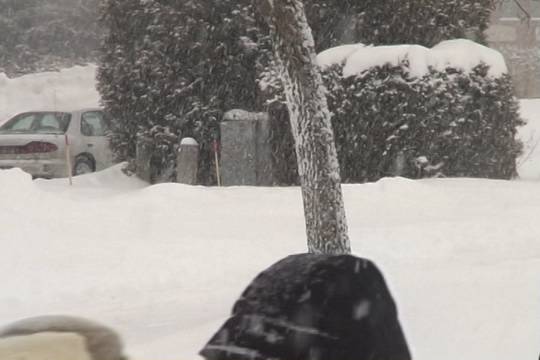}
\end{minipage}
\begin{minipage}[b]{0.13\textwidth}
\includegraphics[width=2.3cm,height=1.7cm]{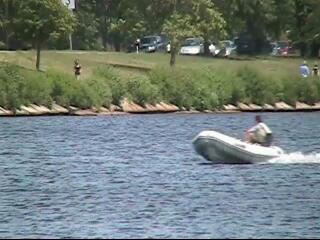}
\end{minipage}
\begin{minipage}[b]{0.13\textwidth}
\includegraphics[width=2.3cm,height=1.7cm]{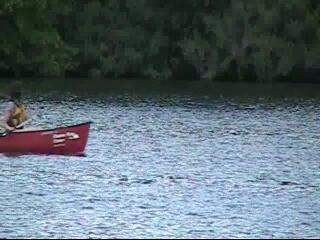}
\end{minipage}
\begin{minipage}[b]{0.13\textwidth}
\includegraphics[width=2.3cm,height=1.7cm]{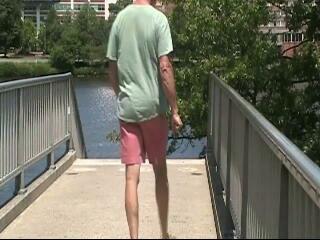}
\end{minipage}
\begin{minipage}[b]{0.13\textwidth}
\includegraphics[width=2.3cm,height=1.7cm]{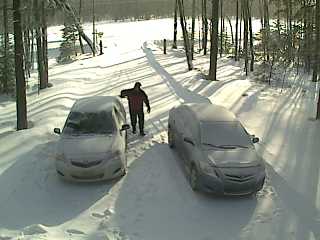}
\end{minipage}
}
\subfigure{
\begin{minipage}[b]{0.13\textwidth}
\includegraphics[width=2.3cm,height=1.7cm]{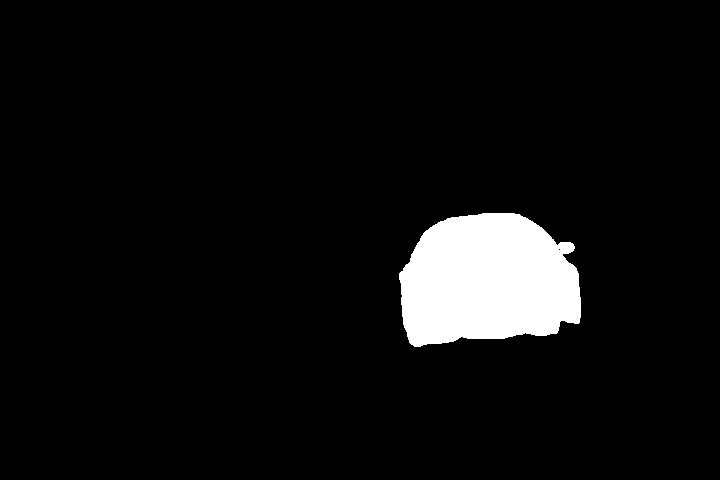}
\end{minipage}
\begin{minipage}[b]{0.13\textwidth}
\includegraphics[width=2.3cm,height=1.7cm]{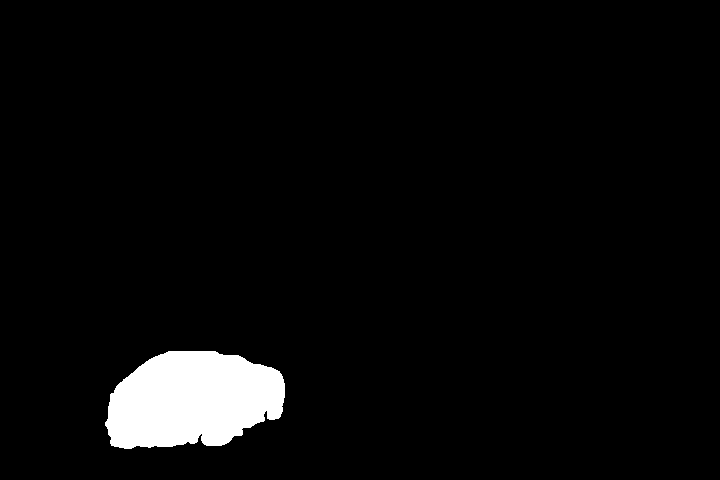}
\end{minipage}
\begin{minipage}[b]{0.13\textwidth}
\includegraphics[width=2.3cm,height=1.7cm]{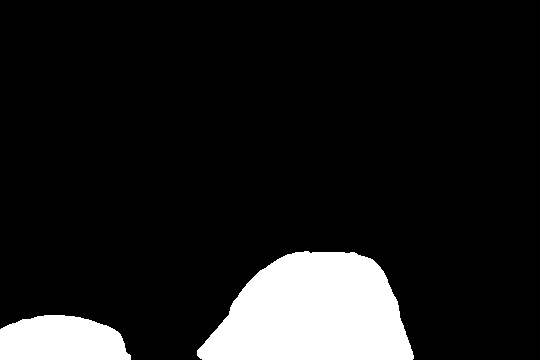}
\end{minipage}
\begin{minipage}[b]{0.13\textwidth}
\includegraphics[width=2.3cm,height=1.7cm]{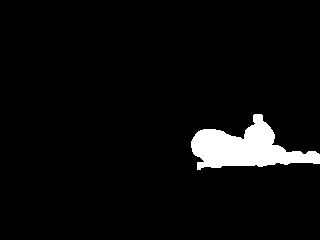}
\end{minipage}
\begin{minipage}[b]{0.13\textwidth}
\includegraphics[width=2.3cm,height=1.7cm]{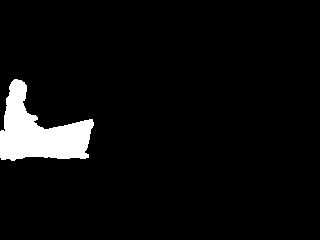}
\end{minipage}
\begin{minipage}[b]{0.13\textwidth}
\includegraphics[width=2.3cm,height=1.7cm]{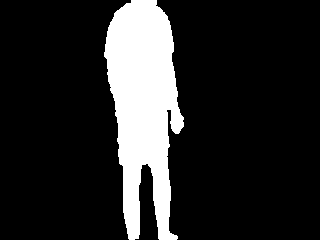}
\end{minipage}
\begin{minipage}[b]{0.13\textwidth}
\includegraphics[width=2.3cm,height=1.7cm]{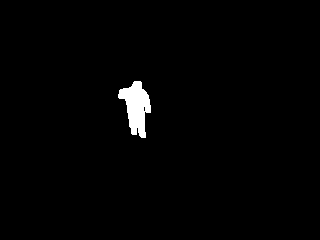}
\end{minipage}
}
\subfigure{
\begin{minipage}[b]{0.13\textwidth}
\includegraphics[width=2.3cm,height=1.7cm]{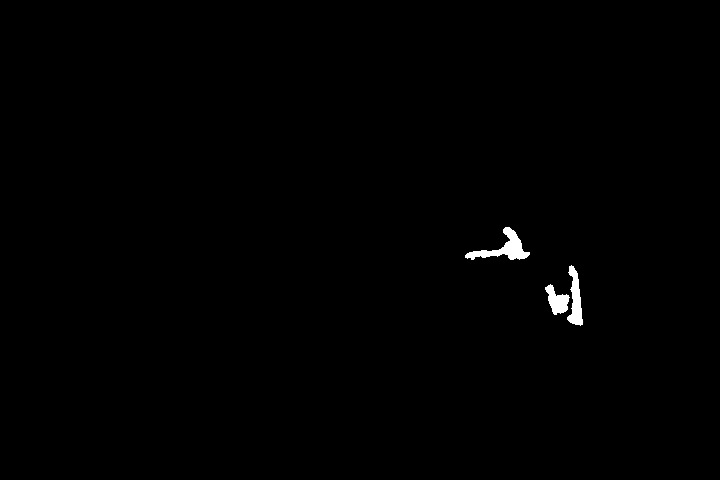}
\end{minipage}
\begin{minipage}[b]{0.13\textwidth}
\includegraphics[width=2.3cm,height=1.7cm]{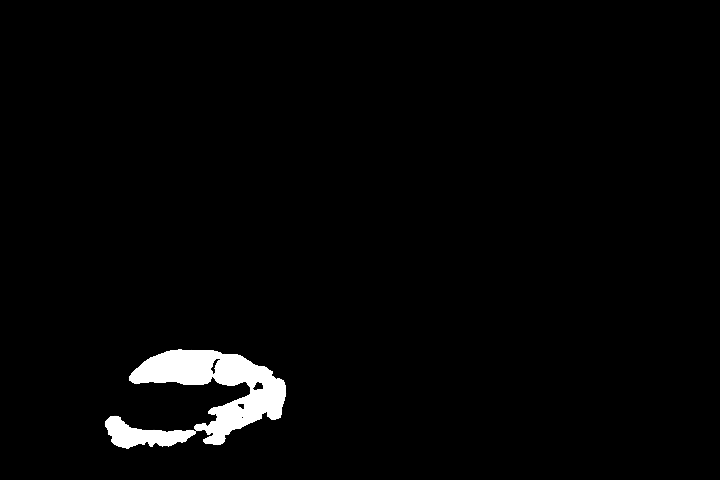}
\end{minipage}
\begin{minipage}[b]{0.13\textwidth}
\includegraphics[width=2.3cm,height=1.7cm]{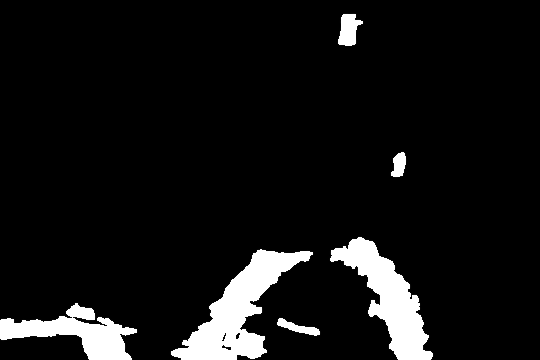}
\end{minipage}
\begin{minipage}[b]{0.13\textwidth}
\includegraphics[width=2.3cm,height=1.7cm]{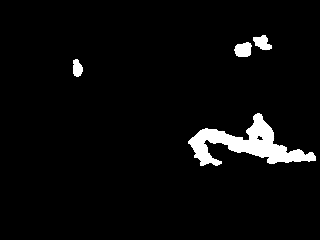}
\end{minipage}
\begin{minipage}[b]{0.13\textwidth}
\includegraphics[width=2.3cm,height=1.7cm]{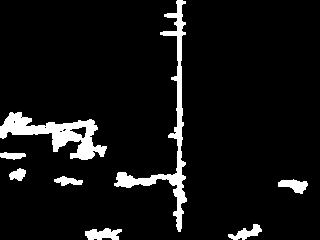}
\end{minipage}
\begin{minipage}[b]{0.13\textwidth}
\includegraphics[width=2.3cm,height=1.7cm]{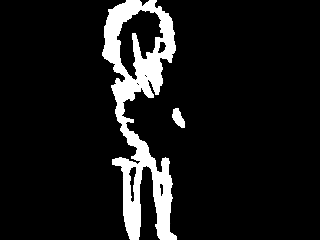}
\end{minipage}
\begin{minipage}[b]{0.13\textwidth}
\includegraphics[width=2.3cm,height=1.7cm]{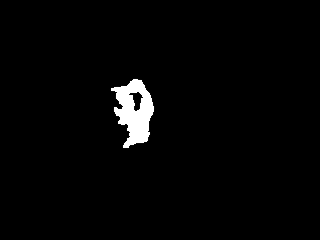}
\end{minipage}
}
\subfigure{
\begin{minipage}[b]{0.13\textwidth}
\includegraphics[width=2.3cm,height=1.7cm]{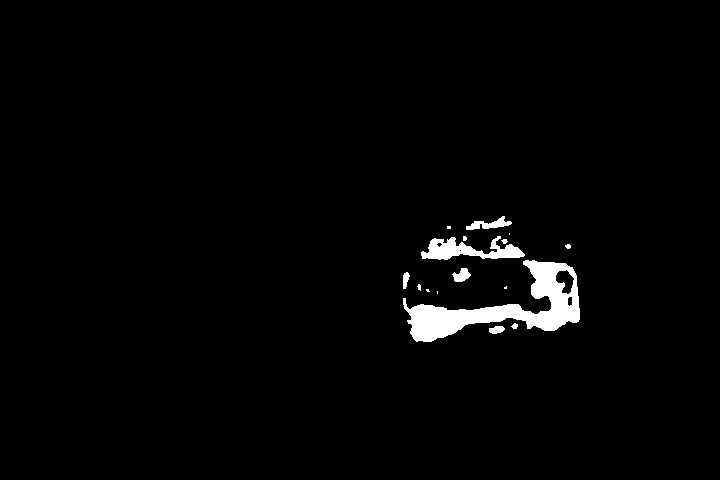}
\end{minipage}
\begin{minipage}[b]{0.13\textwidth}
\includegraphics[width=2.3cm,height=1.7cm]{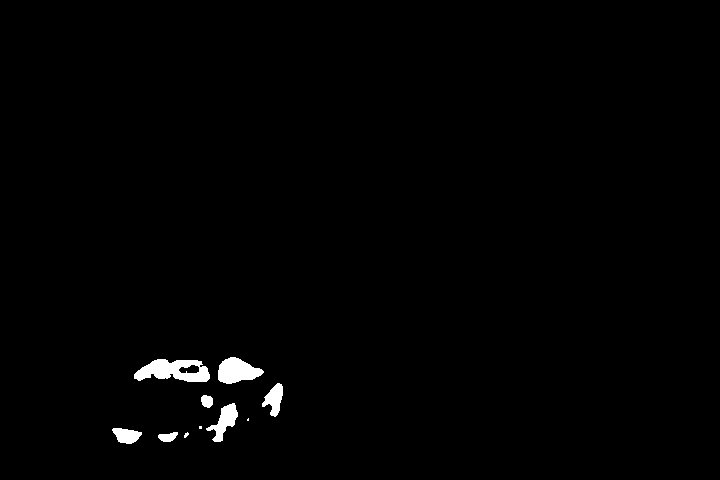}
\end{minipage}
\begin{minipage}[b]{0.13\textwidth}
\includegraphics[width=2.3cm,height=1.7cm]{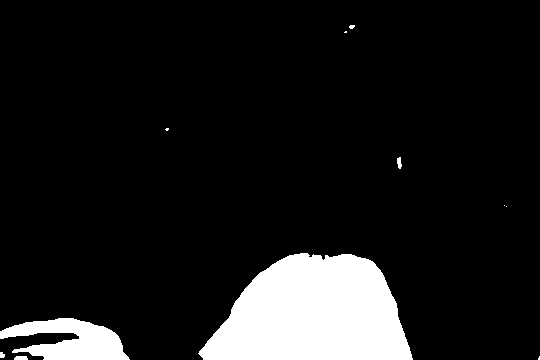}
\end{minipage}
\begin{minipage}[b]{0.13\textwidth}
\includegraphics[width=2.3cm,height=1.7cm]{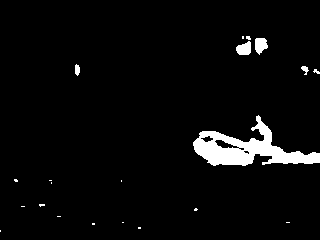}
\end{minipage}
\begin{minipage}[b]{0.13\textwidth}
\includegraphics[width=2.3cm,height=1.7cm]{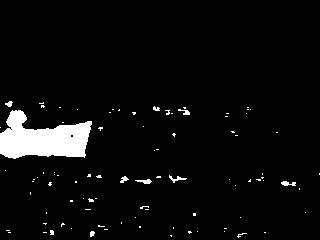}
\end{minipage}
\begin{minipage}[b]{0.13\textwidth}
\includegraphics[width=2.3cm,height=1.7cm]{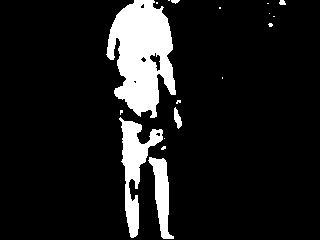}
\end{minipage}
\begin{minipage}[b]{0.13\textwidth}
\includegraphics[width=2.3cm,height=1.7cm]{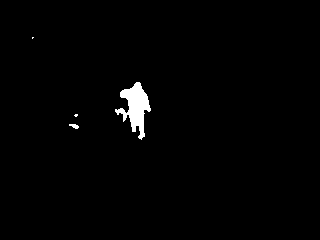}
\end{minipage}
}
\subfigure{
\begin{minipage}[b]{0.13\textwidth}
\includegraphics[width=2.3cm,height=1.7cm]{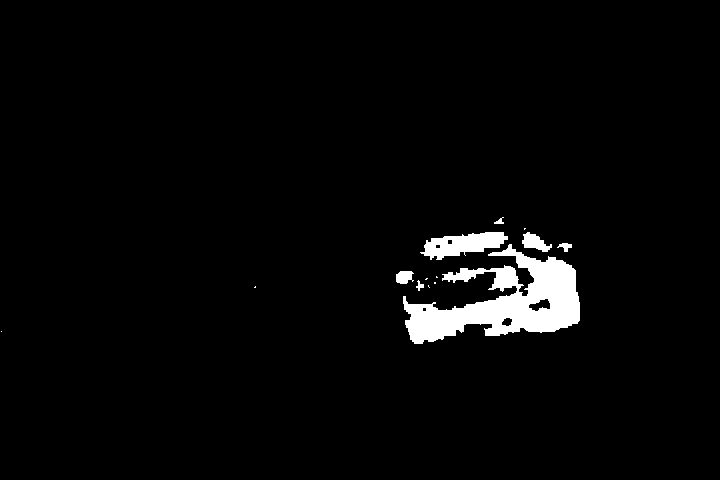}
\end{minipage}
\begin{minipage}[b]{0.13\textwidth}
\includegraphics[width=2.3cm,height=1.7cm]{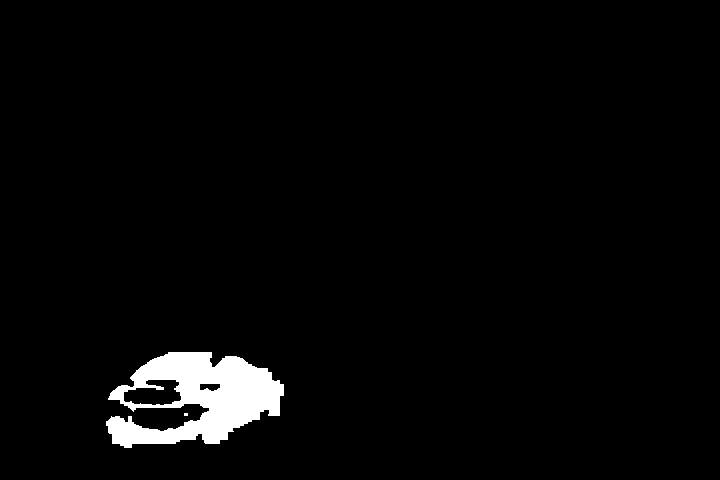}
\end{minipage}
\begin{minipage}[b]{0.13\textwidth}
\includegraphics[width=2.3cm,height=1.7cm]{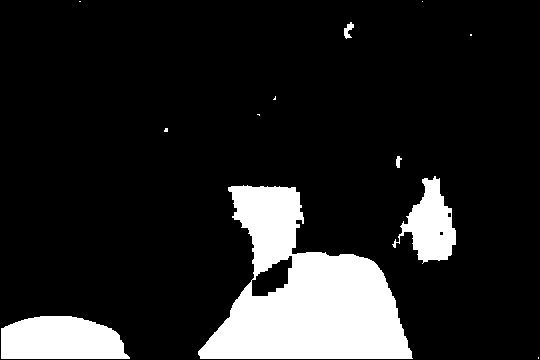}
\end{minipage}
\begin{minipage}[b]{0.13\textwidth}
\includegraphics[width=2.3cm,height=1.7cm]{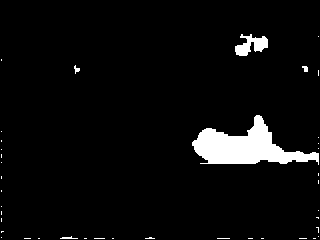}
\end{minipage}
\begin{minipage}[b]{0.13\textwidth}
\includegraphics[width=2.3cm,height=1.7cm]{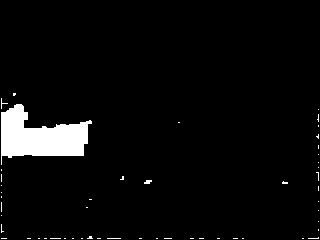}
\end{minipage}
\begin{minipage}[b]{0.13\textwidth}
\includegraphics[width=2.3cm,height=1.7cm]{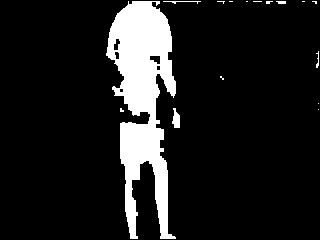}
\end{minipage}
\begin{minipage}[b]{0.13\textwidth}
\includegraphics[width=2.3cm,height=1.7cm]{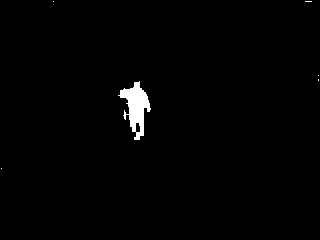}
\end{minipage}
}
\subfigure{
\begin{minipage}[b]{0.13\textwidth}
\includegraphics[width=2.3cm,height=1.7cm]{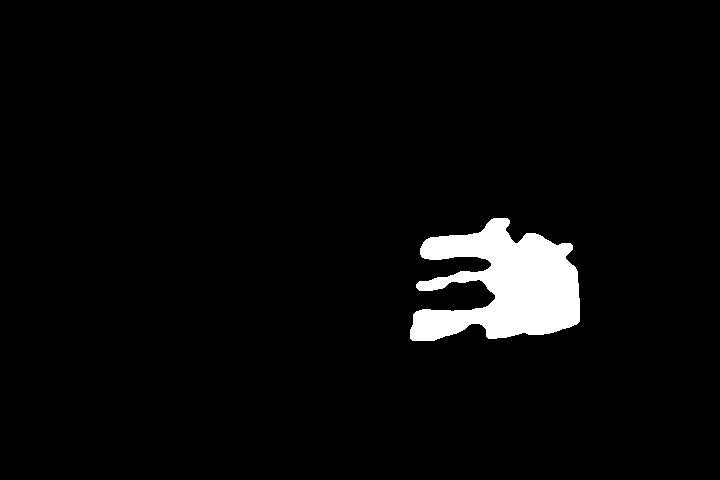}
\end{minipage}
\begin{minipage}[b]{0.13\textwidth}
\includegraphics[width=2.3cm,height=1.7cm]{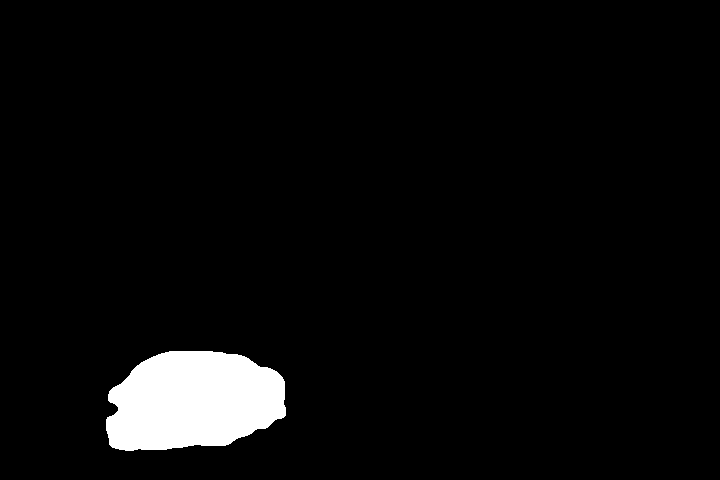}
\end{minipage}
\begin{minipage}[b]{0.13\textwidth}
\includegraphics[width=2.3cm,height=1.7cm]{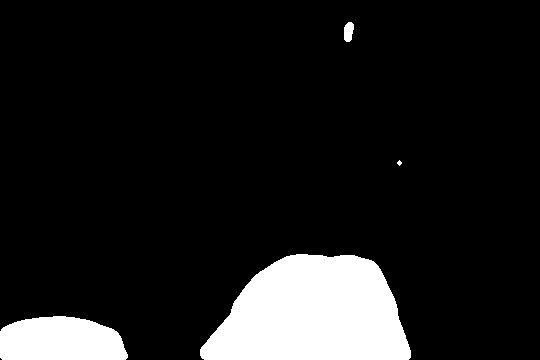}
\end{minipage}
\begin{minipage}[b]{0.13\textwidth}
\includegraphics[width=2.3cm,height=1.7cm]{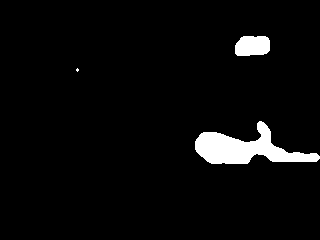}
\end{minipage}
\begin{minipage}[b]{0.13\textwidth}
\includegraphics[width=2.3cm,height=1.7cm]{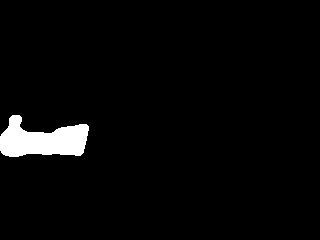}
\end{minipage}
\begin{minipage}[b]{0.13\textwidth}
\includegraphics[width=2.3cm,height=1.7cm]{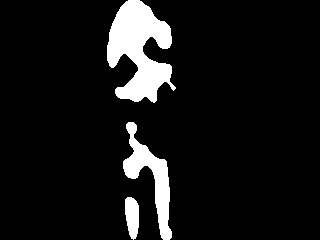}
\end{minipage}
\begin{minipage}[b]{0.13\textwidth}
\includegraphics[width=2.3cm,height=1.7cm]{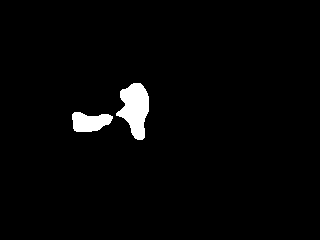}
\end{minipage}
}
\subfigure{
\begin{minipage}[b]{0.13\textwidth}
\includegraphics[width=2.3cm,height=1.7cm]{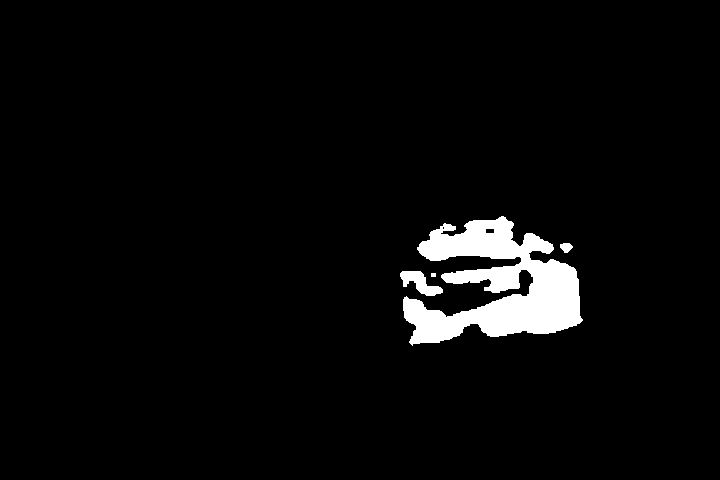}
\end{minipage}
\begin{minipage}[b]{0.13\textwidth}
\includegraphics[width=2.3cm,height=1.7cm]{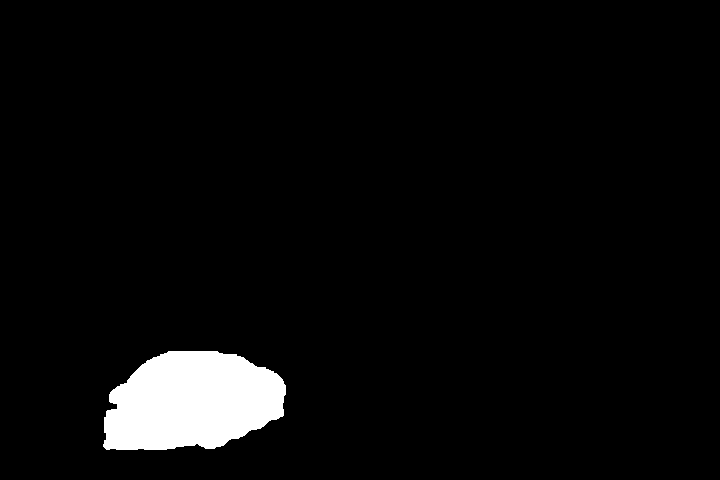}
\end{minipage}
\begin{minipage}[b]{0.13\textwidth}
\includegraphics[width=2.3cm,height=1.7cm]{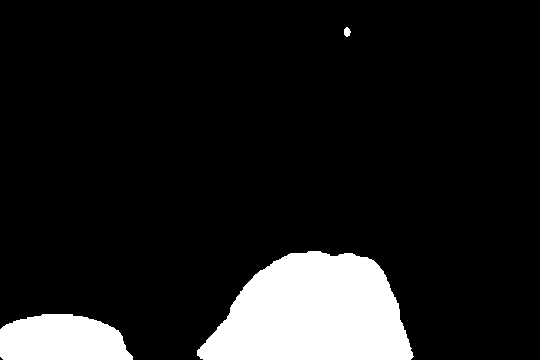}
\end{minipage}
\begin{minipage}[b]{0.13\textwidth}
\includegraphics[width=2.3cm,height=1.7cm]{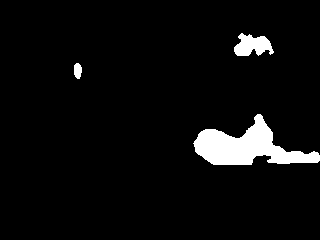}
\end{minipage}
\begin{minipage}[b]{0.13\textwidth}
\includegraphics[width=2.3cm,height=1.7cm]{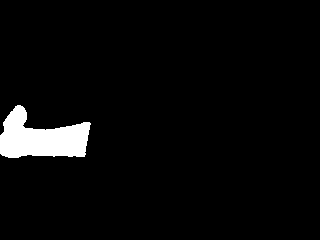}
\end{minipage}
\begin{minipage}[b]{0.13\textwidth}
\includegraphics[width=2.3cm,height=1.7cm]{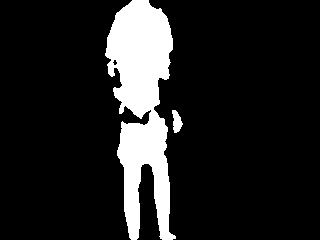}
\end{minipage}
\begin{minipage}[b]{0.13\textwidth}
\includegraphics[width=2.3cm,height=1.7cm]{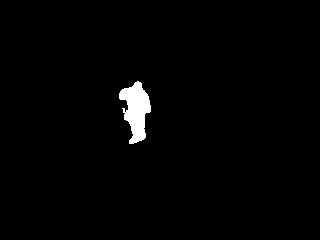}
\end{minipage}
}
\subfigure{
\begin{minipage}[b]{0.13\textwidth}
\includegraphics[width=2.3cm,height=1.7cm]{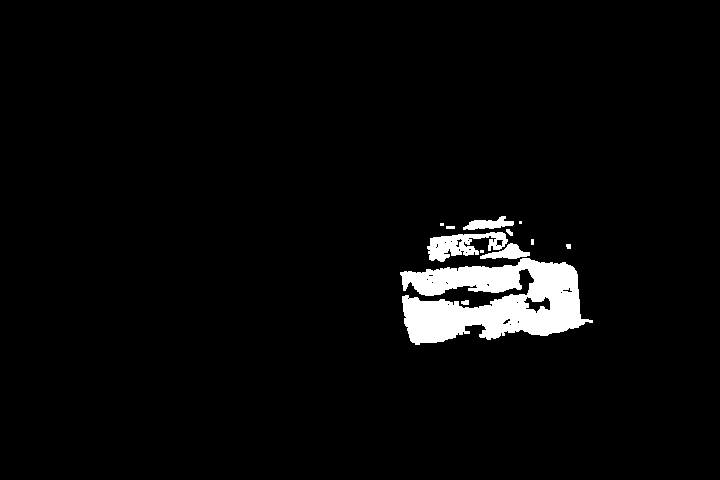}
\end{minipage}
\begin{minipage}[b]{0.13\textwidth}
\includegraphics[width=2.3cm,height=1.7cm]{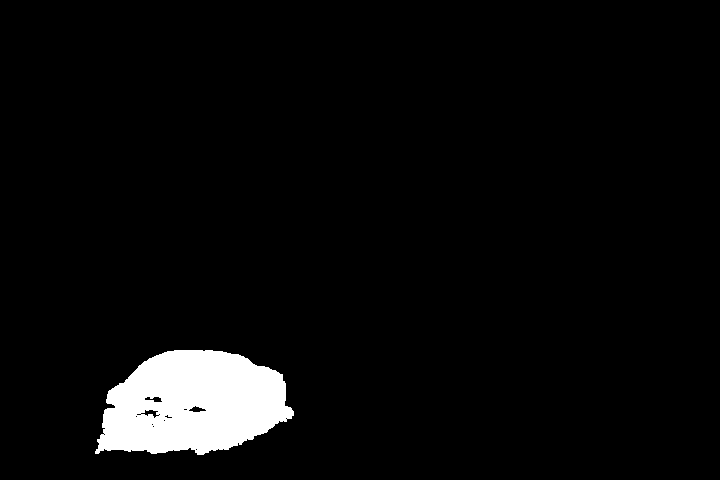}
\end{minipage}
\begin{minipage}[b]{0.13\textwidth}
\includegraphics[width=2.3cm,height=1.7cm]{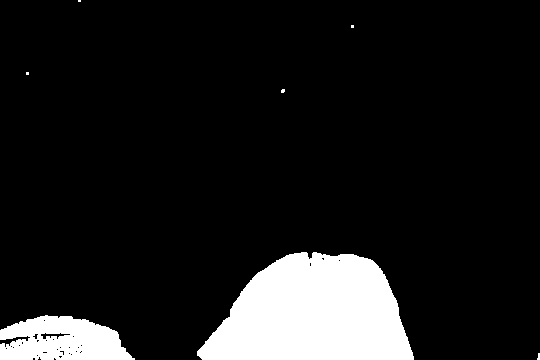}
\end{minipage}
\begin{minipage}[b]{0.13\textwidth}
\includegraphics[width=2.3cm,height=1.7cm]{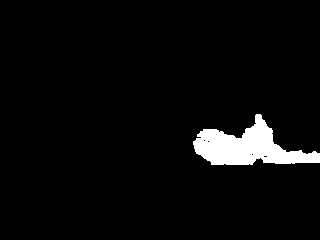}
\end{minipage}
\begin{minipage}[b]{0.13\textwidth}
\includegraphics[width=2.3cm,height=1.7cm]{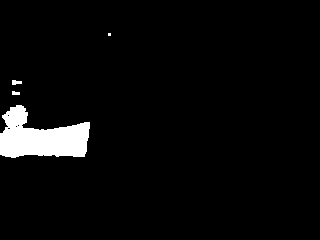}
\end{minipage}
\begin{minipage}[b]{0.13\textwidth}
\includegraphics[width=2.3cm,height=1.7cm]{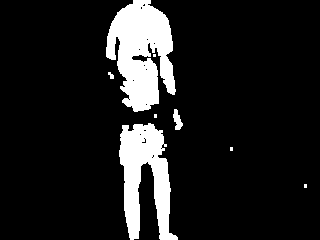}
\end{minipage}
\begin{minipage}[b]{0.13\textwidth}
\includegraphics[width=2.3cm,height=1.7cm]{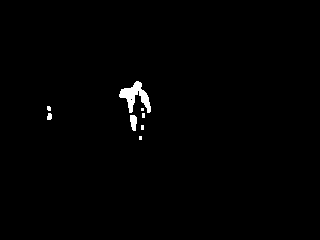}
\end{minipage}
}
\subfigure{
\begin{minipage}[b]{0.13\textwidth}
\includegraphics[width=2.3cm,height=1.7cm]{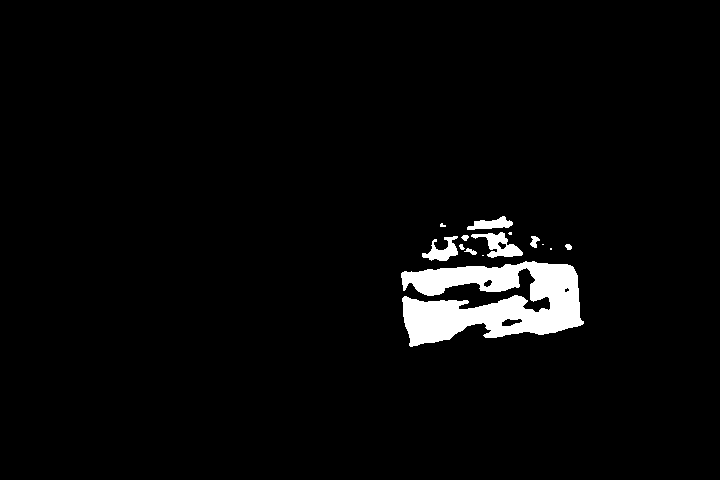}
\end{minipage}
\begin{minipage}[b]{0.13\textwidth}
\includegraphics[width=2.3cm,height=1.7cm]{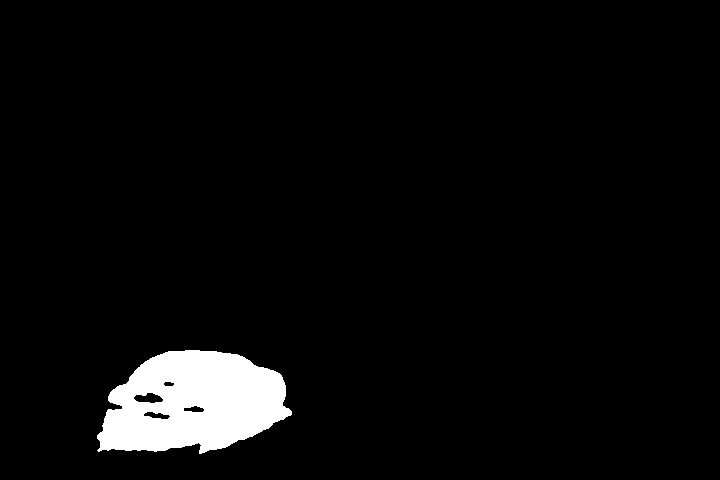}
\end{minipage}
\begin{minipage}[b]{0.13\textwidth}
\includegraphics[width=2.3cm,height=1.7cm]{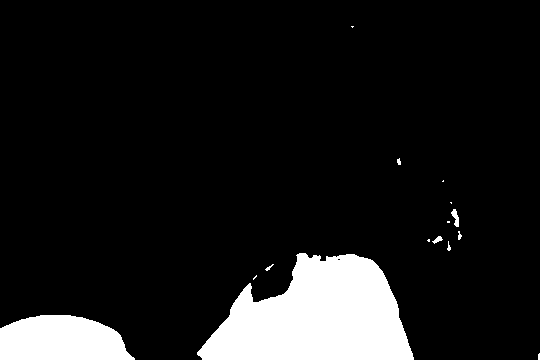}
\end{minipage}
\begin{minipage}[b]{0.13\textwidth}
\includegraphics[width=2.3cm,height=1.7cm]{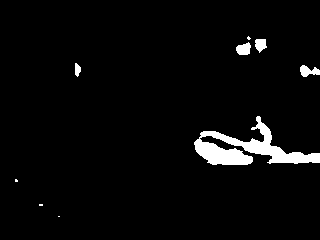}
\end{minipage}
\begin{minipage}[b]{0.13\textwidth}
\includegraphics[width=2.3cm,height=1.7cm]{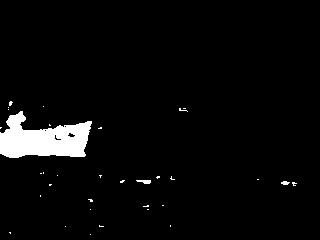}
\end{minipage}
\begin{minipage}[b]{0.13\textwidth}
\includegraphics[width=2.3cm,height=1.7cm]{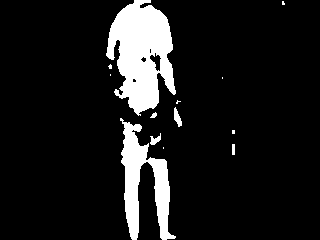}
\end{minipage}
\begin{minipage}[b]{0.13\textwidth}
\includegraphics[width=2.3cm,height=1.7cm]{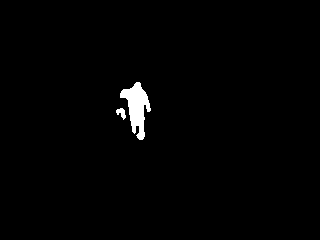}
\end{minipage}
}
\subfigure{
\begin{minipage}[b]{0.13\textwidth}
\includegraphics[width=2.3cm,height=1.7cm]{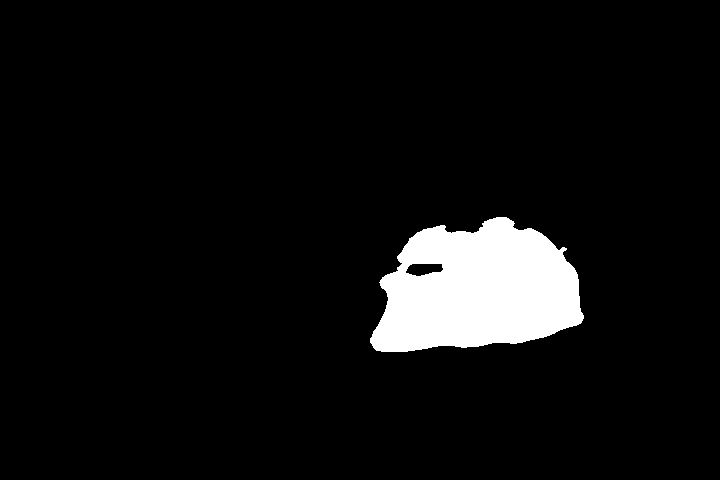}
\end{minipage}
\begin{minipage}[b]{0.13\textwidth}
\includegraphics[width=2.3cm,height=1.7cm]{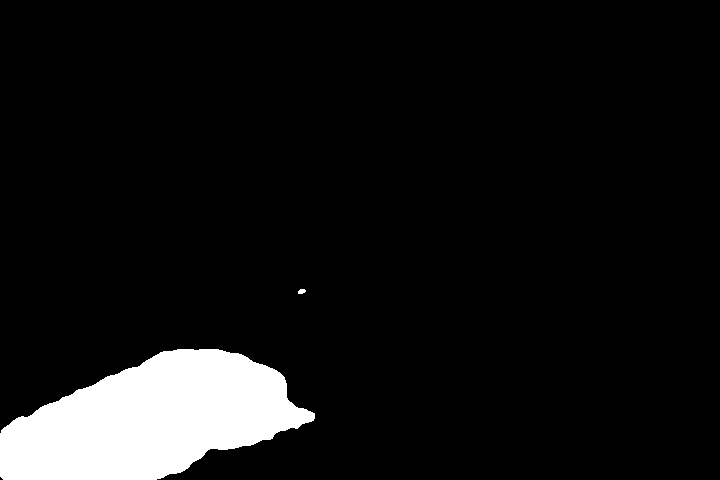}
\end{minipage}
\begin{minipage}[b]{0.13\textwidth}
\includegraphics[width=2.3cm,height=1.7cm]{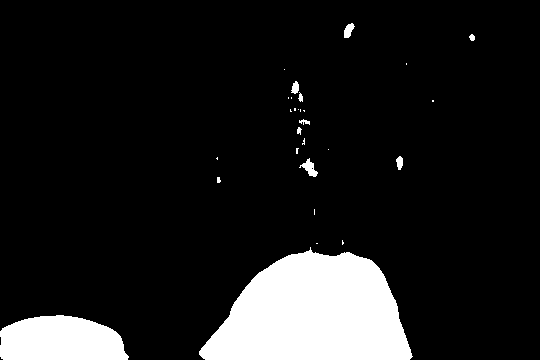}
\end{minipage}
\begin{minipage}[b]{0.13\textwidth}
\includegraphics[width=2.3cm,height=1.7cm]{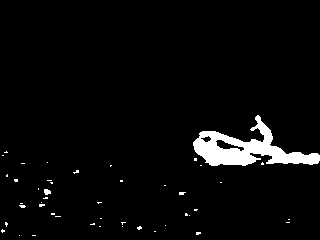}
\end{minipage}
\begin{minipage}[b]{0.13\textwidth}
\includegraphics[width=2.3cm,height=1.7cm]{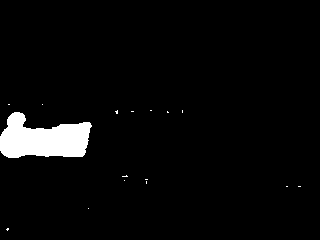}
\end{minipage}
\begin{minipage}[b]{0.13\textwidth}
\includegraphics[width=2.3cm,height=1.7cm]{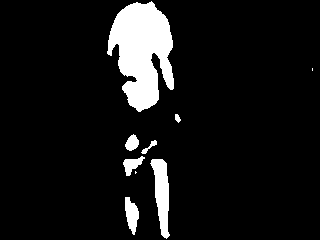}
\end{minipage}
\begin{minipage}[b]{0.13\textwidth}
\includegraphics[width=2.3cm,height=1.7cm]{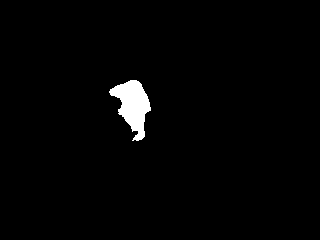}
\end{minipage}
}
\subfigure{
\begin{minipage}[b]{0.13\textwidth}
\includegraphics[width=2.3cm,height=1.7cm]{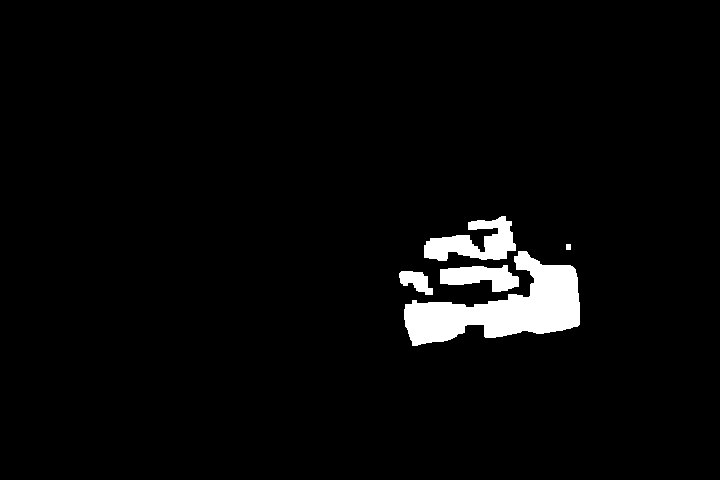}
\end{minipage}
\begin{minipage}[b]{0.13\textwidth}
\includegraphics[width=2.3cm,height=1.7cm]{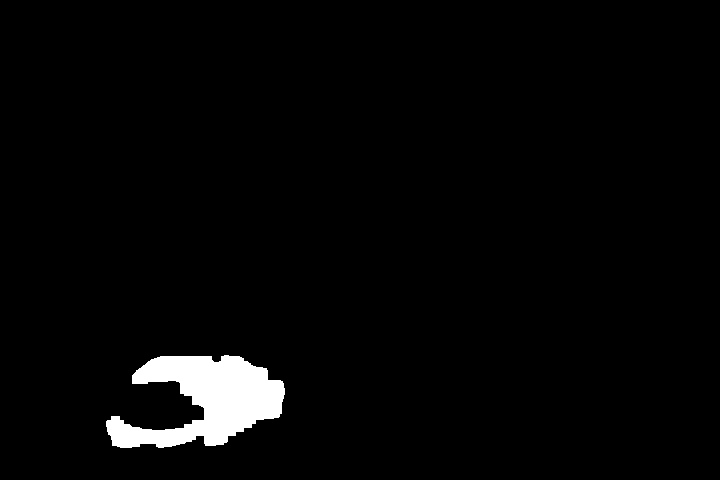}
\end{minipage}
\begin{minipage}[b]{0.13\textwidth}
\includegraphics[width=2.3cm,height=1.7cm]{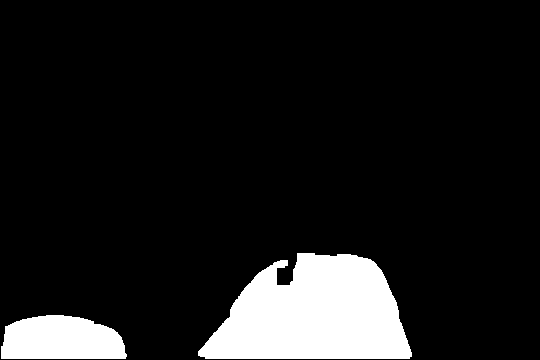}
\end{minipage}
\begin{minipage}[b]{0.13\textwidth}
\includegraphics[width=2.3cm,height=1.7cm]{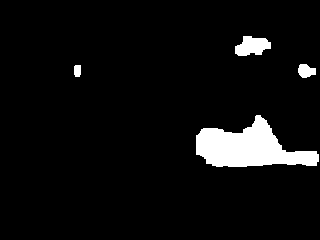}
\end{minipage}
\begin{minipage}[b]{0.13\textwidth}
\includegraphics[width=2.3cm,height=1.7cm]{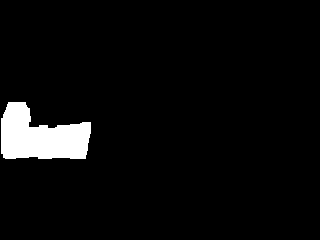}
\end{minipage}
\begin{minipage}[b]{0.13\textwidth}
\includegraphics[width=2.3cm,height=1.7cm]{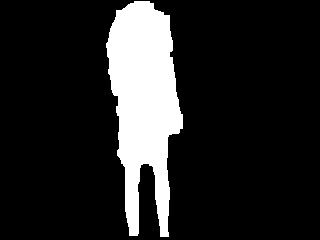}
\end{minipage}
\begin{minipage}[b]{0.13\textwidth}
\includegraphics[width=2.3cm,height=1.7cm]{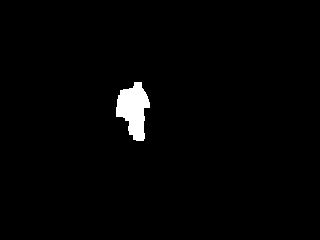}
\end{minipage}
}
\subfigure{
\begin{minipage}[b]{0.13\textwidth}
\includegraphics[width=2.3cm,height=1.7cm]{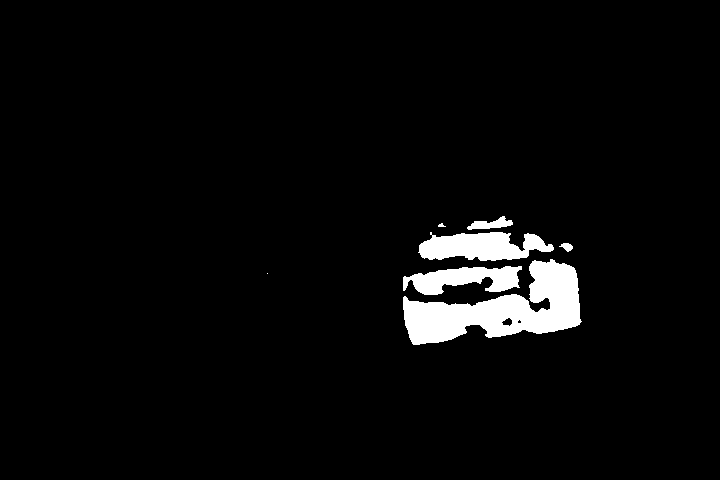}
\end{minipage}
\begin{minipage}[b]{0.13\textwidth}
\includegraphics[width=2.3cm,height=1.7cm]{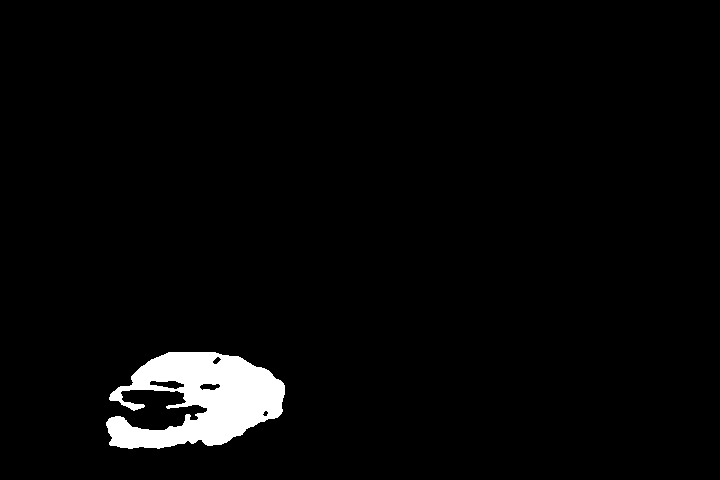}
\end{minipage}
\begin{minipage}[b]{0.13\textwidth}
\includegraphics[width=2.3cm,height=1.7cm]{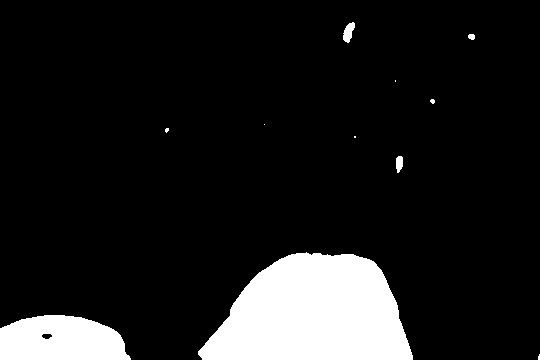}
\end{minipage}
\begin{minipage}[b]{0.13\textwidth}
\includegraphics[width=2.3cm,height=1.7cm]{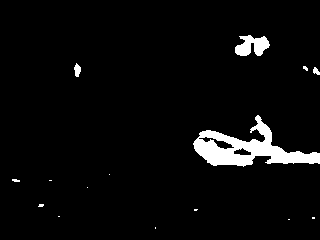}
\end{minipage}
\begin{minipage}[b]{0.13\textwidth}
\includegraphics[width=2.3cm,height=1.7cm]{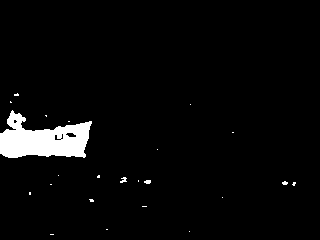}
\end{minipage}
\begin{minipage}[b]{0.13\textwidth}
\includegraphics[width=2.3cm,height=1.7cm]{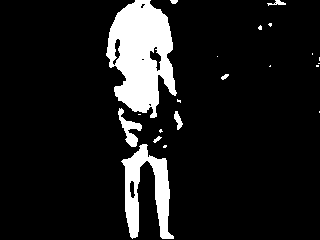}
\end{minipage}
\begin{minipage}[b]{0.13\textwidth}
\includegraphics[width=2.3cm,height=1.7cm]{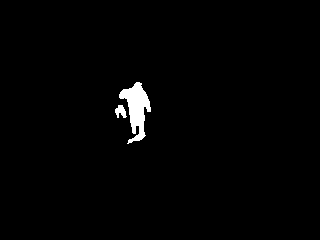}
\end{minipage}
}
\subfigure{
\begin{minipage}[b]{0.13\textwidth}
\includegraphics[width=2.3cm,height=1.7cm]{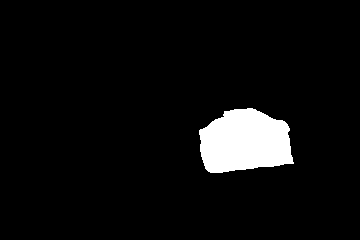}
\end{minipage}
\begin{minipage}[b]{0.13\textwidth}
\includegraphics[width=2.3cm,height=1.7cm]{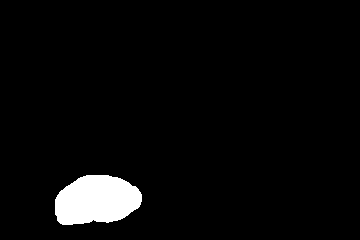}
\end{minipage}
\begin{minipage}[b]{0.13\textwidth}
\includegraphics[width=2.3cm,height=1.7cm]{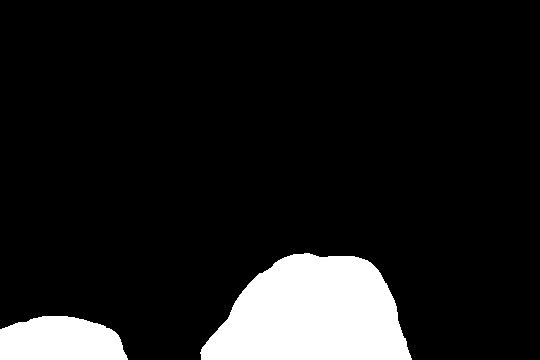}
\end{minipage}
\begin{minipage}[b]{0.13\textwidth}
\includegraphics[width=2.3cm,height=1.7cm]{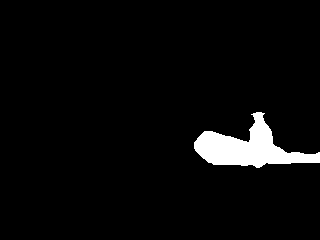}
\end{minipage}
\begin{minipage}[b]{0.13\textwidth}
\includegraphics[width=2.3cm,height=1.7cm]{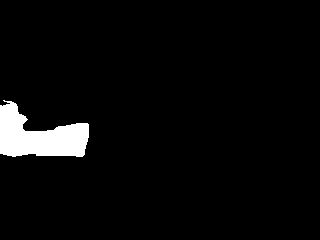}
\end{minipage}
\begin{minipage}[b]{0.13\textwidth}
\includegraphics[width=2.3cm,height=1.7cm]{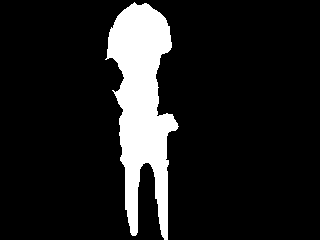}
\end{minipage}
\begin{minipage}[b]{0.13\textwidth}
\includegraphics[width=2.3cm,height=1.7cm]{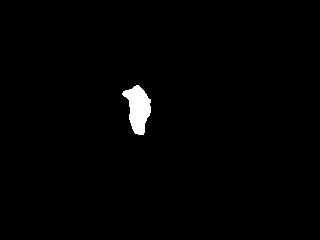}
\end{minipage}
}
\caption{Some examples of the detection results produced by eleven methods. From top to bottom: original images, ground truth, XUE\cite{paper17}, Euclidean distance\cite{paperEuclideandistance}, MST\cite{paper12}, Spectral\cite{paper11}, FSTG\cite{paper10}, CwisarDH\cite{paper14}, GMM\cite{paperGMM}, CP3online\cite{paper13}, BinWang\cite{paperBinWang}, KDE\cite{paper08}, and the proposed mehtod.}
\label{fig:picture:11res}
\end{figure*}

Table~\ref{tab:performance_comparison} shows evaluation results on seven videos from the CDCNET 2014 database. It can be seen that, except for the overpass and winterDriveway videos, the proposed method achieves a F-measure that is similar to or higher than the competing methods.

Figure~\ref{fig:picture:11res} gives some examples of the detection results, demonstrating the ability of the proposed method in dealing with complex dynamic scenes. Regarding the first column that shows car moving in the snowy weather, only CP3_online\cite{paper13} and the proposed method perform well, while Xue\cite{paper17} and MST\cite{paper12} lost many foregrounds. In the second column that also shows car moving in the snowy weather (but the weather is better than the first column), Spectral\cite{paper11}, FTSG\cite{paper10}, CwisarDH\cite{paper14}, GMM\cite{paperGMM}, CP3_online\cite{paper13} and the proposed method have good results. While dealing with the third column that shows people skating in the snowy weather, Xue\cite{paper17} loses many foregrounds in the scenes ``Skating'', ``snowfall'' and ``Blizzard''. On the forth and fifth columns that show boats and canoe in dynamic water surface, Xue\cite{paper17} loses many foregrounds in the scenes ``canoe'' and ``boats'', while Euclideandistance\cite{paperEuclideandistance}, GMM\cite{paperGMM}, CP3_online\cite{paper13} and  KDE\cite{paper08} wrongly detect some background as foreground. The sixth column shows a man walking in the scene with swaying leaves, where BinWang\cite{paperBinWang} gets better results. In the seventh column with intermittent motion, GMM\cite{paperGMM} loses much foreground.
\section{Conclusion}\label{sec:con}
RPCA is sensitive to dynamic background and bad weather. To overcome this problem, firstly, instead of using the raw pixel-value as features that are brittle while applying to dynamic background, a Gaussian max-pooling operator is used to estimate the “stable-value” for each pixel. Those stable-pixels are robust to various background. Then, the video segment is used to incorporate the temporal and spatial continuity of both the foreground and background for group sparsity constrain. The proposed method is tested in some challenging situations. Compared with other popular methods, the experimental results demonstrate the proposed has better performance for the dynamic background and bad weather.
\section*{Acknowledgment}
The work of Guangcan Liu is supported in part by National Natural Science Foundation of China (NSFC) under Grant 61622305 and Grant 61502238, in part by the Natural Science Foundation of Jiangsu Province of China (NSFJPC) under Grant BK20160040.
\ifCLASSOPTIONcaptionsoff
  \newpage
\fi

\bibliographystyle{IEEEtran}
\bibliography{bare_jrnl}

\begin{IEEEbiography}[{\includegraphics[width=1in,
height=1.25in,clip, keepaspectratio]{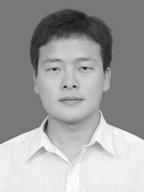}}]{Yang Li} received the bachelor's degree from Nanjing University of Information Science and Technology, Nanjing, China, in 2013, and the master's degree from Beifang Univesity of Nationalities, Yinchuan, China, in 2016. Now he is a PhD candidate in Tianjin University of Technology. His research interests touch on the areas of computer vision, and image processing.
\end{IEEEbiography}

\begin{IEEEbiography}[{\includegraphics[width=1in,
height=1.25in,clip, keepaspectratio]{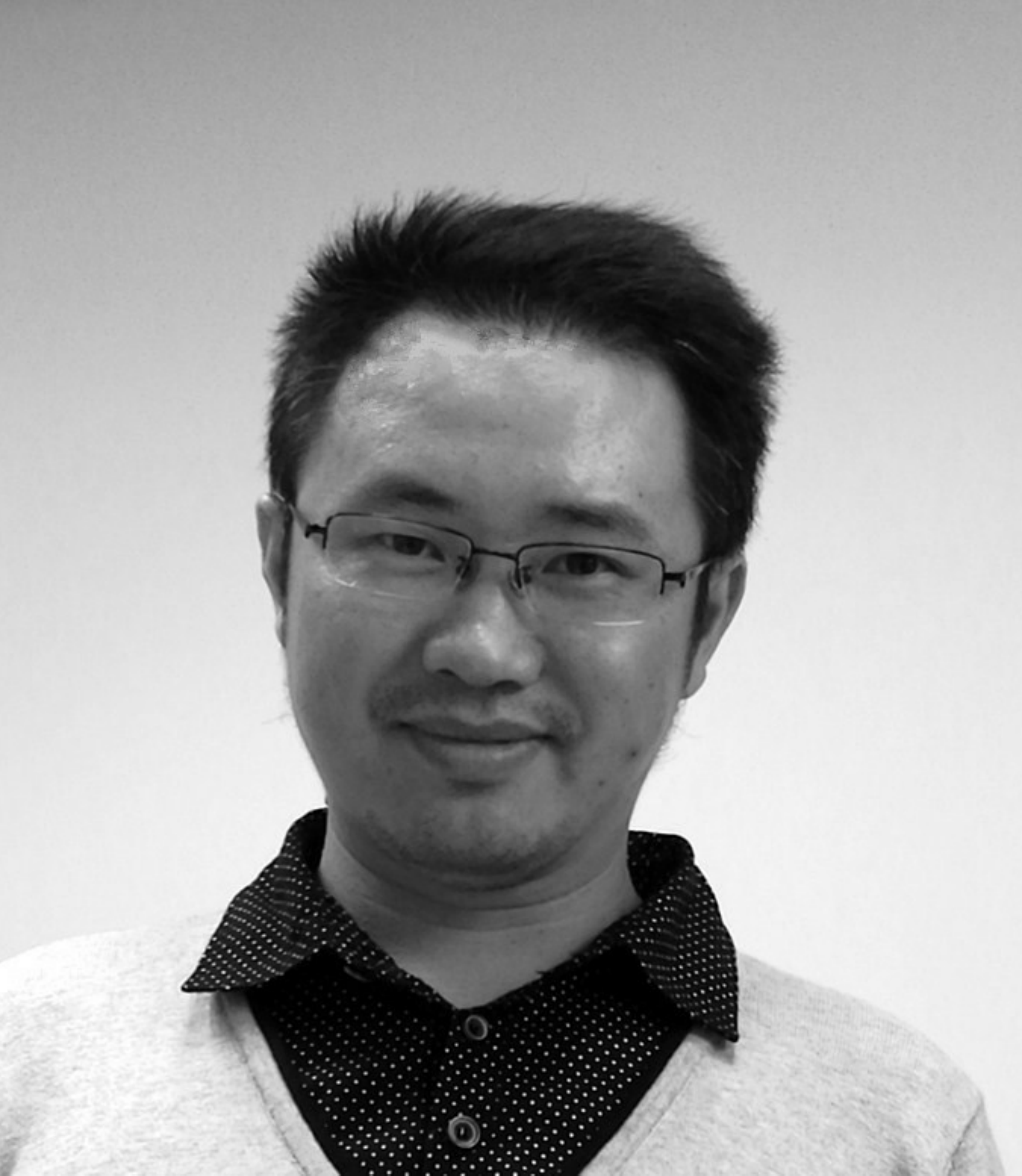}}]{Guangcan Liu}(M'11) received the bachelor's degree in mathematics and the Ph.D. degree in computer science and engineering from Shanghai Jiao Tong University, Shanghai, China, in 2004 and 2010, respectively. He was a Post-Doctoral Researcher with the National University of Singapore, Singapore, from 2011 to 2012, the University of Illinois at Urbana-Champaign, Champaign, IL, USA, from 2012 to 2013, Cornell University, Ithaca, NY, USA, from 2013 to 2014, and Rutgers University, Piscataway, NJ, USA, in 2014. Since 2014, he has been a Professor with the School of Information and Control, Nanjing University of Information Science and Technology, Nanjing, China. His research interests touch on the areas of machine learning, computer vision, and image processing.
\end{IEEEbiography}

\begin{IEEEbiography}[{\includegraphics[width=1in,
height=1.25in,clip, keepaspectratio]{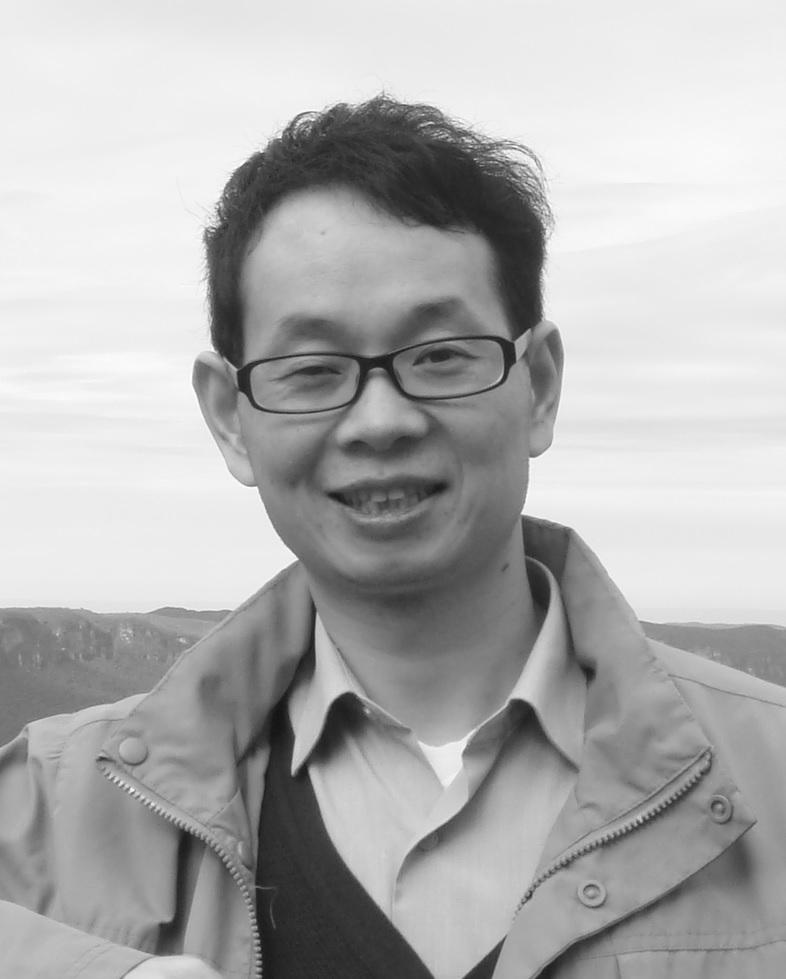}}]{Shengyong Chen}(M'01 - SM'10) Shengyong Chen received the Ph.D. degree in computer vision from City University of Hong Kong, Hong Kong, in 2003. He is currently a Professor of Tianjin University of Technology and Zhejiang University of Technology, China. He received a fellowship from the Alexander von Humboldt Foundation of Germany and worked at University of Hamburg in 2006 - 2007. His research interests include computer vision, robotics, and image analysis. Dr. Chen is a Fellow of IET and senior member of IEEE and CCF. He has published over 100 scientific papers in international journals. He received the National Outstanding Youth Foundation Award of China in 2013.
\end{IEEEbiography}

\end{document}